\documentclass[conference]{IEEEtran}
\usepackage{xcolor}
\definecolor{darkgray}{gray}{0.35}
\usepackage{tikz}
\usetikzlibrary{bayesnet}
\usepackage{algorithm}
\usepackage{algpseudocode}
\usepackage{color, soul}
\usepackage{setspace}
\usepackage{placeins}
\usepackage{graphicx}
\usepackage{caption}
\usepackage{subfigure}
\usepackage{cite}
\usepackage{amsmath}
\usepackage{url}
\newtheorem{definition}{Definition}
\renewcommand{\baselinestretch}{0.872}
\begin{document}
\bstctlcite{IEEE:nodash}
\title{Source-LDA: Enhancing probabilistic topic models using prior knowledge sources}
\author{
\IEEEauthorblockN{Justin Wood$^{1\text{,}2}$, Patrick Tan$^1$, Wei Wang$^1$, Corey Arnold$^2$}
\IEEEauthorblockA{$^1$Department of Computer Science, University of California Los Angeles, CA 90095, USA\\ 
$^2$Medical Imaging Informatics Group, University of California, Los Angeles, CA 90095, USA\\
\texttt{\{juwood03, patrickptt, cwarnold\}@ucla.edu}, \texttt{weiwang@cs.ucla.edu}\\  \\}}
\maketitle
\begin{abstract}
Topic modeling has increasingly attracted interests from researchers.
Common methods of topic modeling usually produce a collection of
unlabeled topics where each topic is depicted by a distribution of words.
Associating semantic meaning with these word distributions is not always straightforward.
Traditionally, this task is left to human interpretation.
Manually labeling the topics is unfortunately not always easy,
as topics generated by unsupervised learning methods do not necessarily
align well with our prior knowledge in the subject domains. Currently, two approaches
to solve this issue exist. The first is a post-processing procedure
that assigns each topic with a label from the prior knowledge base that is
semantically closest to the word distribution of the topic.
The second is a supervised topic modeling approach that restricts the topics
to a predefined set whose word distributions are provided beforehand.
Neither approach is ideal, as the former may produce labels that do not
accurately describe the word distributions, and the latter lacks the ability to detect
unknown topics that are crucial to enrich our knowledge base.
Our goal in this paper is to introduce a
semi-supervised Latent Dirichlet allocation (LDA) model, Source-LDA,
which incorporates prior knowledge to guide the topic modeling process to improve
both the quality of the resulting topics and of the topic labeling.
We accomplish this by integrating existing labeled knowledge sources representing known
potential topics into a probabilistic topic model. These knowledge sources are
translated into a distribution and used to set the hyperparameters of the
Dirichlet generated distribution over words.
This approach ensures that the topic inference process is consistent with existing knowledge,
and simultaneously, allows for discovery of new topics.
The results show improved topic generation and increased accuracy in topic labeling
when compared to those obtained using various labeling approaches based off LDA.
\end{abstract}

\section{Introduction}
Existing topic modeling is often based off
Latent Dirichlet allocation (LDA)~\cite{Blei03latentdirichlet}  and involves analyzing a given
corpus to produce a distribution over words for each latent topic
and a distribution over latent topics for each document. 
The distributions representing topics are often useful and
generally representative of a linguistic topic. 
Unfortunately, assigning labels to these topics is often left to manual interpretation.\par
\indent
Identifying topic labels
is useful in summarizing a set of words
with a single label. For example, words such as pencil,
laptop, ruler, eraser, and book can be mapped to the label
``School Supplies.'' Adding descriptive semantics to each topic can help people,
especially those without domain knowledge, to understand topics
obtained by topic modeling. \par 
\indent
A motivating application of accurate topic labeling
is to develop summarization systems for primary care physicians, who
are faced with the challenges of being inundated with too much data for a 
patient and too little time to comprehend it all~\cite{emr_time}. The labels
can be used to more appropriately and quickly give an overview, or a summary, of patient's medical
history, leading to better outcomes for the patient. This added information can bring significant
value to the field of clinical informatics which
already utilizes topic modeling without
labeling~\cite{arnold2010clinical, BisginLFXT11, corey_meta}.
\par
\indent
Existing approaches in labeling topics usually
do their fitting of labels to
topics after completion of the
unsupervised topic modeling process.
A topic produced by this approach may not
always match well with any semantic concepts and 
would therefore be difficult to categorize with a
single label. 
These problems are best illustrated via a simple case
study.  
\vspace{1mm}
\subsubsection {Case Study}
Suppose a corpus of a news source that consists of two
articles is given by documents $d1$ and $d2$ each with three words: \par
\vspace{1.5mm}
\indent $\mathbf{d1}$ - pencil, pencil, umpire \par
\indent$\mathbf{d2}$ - ruler, ruler, baseball\par
\vspace{1.5mm}
\noindent LDA (with the traditionally used collapsed Gibbs sampler, standard
hyperparameters and the number of topics ($K$) set as two) would
output different results for different runs due to
the inherent stochastic nature. It is very possible to
obtain the following result of topic assignments: \par
\vspace{1.5mm}
\indent $\mathbf{d1}$ - $\text{\textcolor{darkgray}{pencil}}^1$, $\text{\textcolor{darkgray}{pencil}}^1$, $\text{umpire}^2$ \par
\indent$\mathbf{d2}$ - $\text{ruler}^2$, $\text{ruler}^2$, $\text{\textcolor{darkgray}{baseball}}^1$ \par
\vspace{1.5mm}
\noindent But these assignments to topics differs from the
ideal solution that involves knowing the context of the
topics in which these words come from. If the topic
modeling was to incorporate prior knowledge about the
topics ``School Supplies'' and ``Baseball'', then a topic
modeling process will more likely generate the ideal topic
assignments of: \par
\vspace{1.5mm}
\indent $\mathbf{d1}$ - $\text{pencil}^2$, $\text{pencil}^2$, $\text{\textcolor{darkgray}{umpire}}^1$\par
\indent$\mathbf{d2}$ - $\text{ruler}^2$, $\text{ruler}^2$, $\text{\textcolor{darkgray}{baseball}}^1$ \par
\vspace{1.5mm}
\noindent
and assign a label of ``School Supplies'' to topic $1$ and
``Baseball'' to topic $2$. Furthermore
it is advantageous to incorporate this
prior knowledge during the topic modeling
process. Consider the following table
displaying four different mapping
techniques of the first result using the Wikipedia articles of ``School Supplies'' and
``Baseball'' as the prior knowledge: 
\begin{table} [!h]
\hspace*{1.5em}\begin{tabular}{ l|l|l}
  Technique & Topic 1 & Topic 2 \\
  \hline %
  JS Divergence & Baseball & Baseball \\
  TF-IDF/CS & (same) & (same) \\
  Counting & Baseball & Baseball \\
  PMI & (same) & (same) \\
\end{tabular}
\end{table}
\par
\noindent 
Applying this labeling post topic
modeling can lead to problems dealing with the
topic themselves. This is not so much a problem
of the mapping techniques but of the
topics used as input. By separating the
topics during inference this problem of combining
different semantic topics can be avoided.\par
\vspace{2mm}
\indent
To overcome this problem, one may take a supervised approach that incorporates such prior knowledge
into the topic modeling process to improve the quality of
topic assignments and more effectively label topics. 
However, existing supervised approaches
~\cite{concept_topic_modeling, explicit, guidedlda} are
either too lenient or too strict. For example, in
the Concept-topic model (CTM)~\cite{concept_topic_modeling}, a multinomial distribution is
placed over known concepts with associated word sets. This
pioneering approach does integrate prior knowledge, but
does not take into account word distributions. For example
if a document is generated about the topic ``School Supplies'' it is much more probable
to see the word ``pencil'' than the word ``compass'' even though
both words may be associated with the topic ``School Supplies''.
This technique also requires some supervision which requires
manually inputting preexisting concepts and their bags of words. \par
\indent
Another approach given by Hansen et al. as explicit Dirichlet allocation~\cite{explicit} incorporates
a preexisting distribution based off Wikipedia but does not allow for variance from the Wikipedia distribution.
This approach fulfills the goal of incorporating prior knowledge with their distributions but requires
the topic in the generated corpus to strictly follow the Wikipedia word distributions.\par
\indent
To address these limitations, we propose the Source-LDA model which is a balance between these two approaches.
The goal is to allow for simultaneous discovery of both known and unknown topics. Given a collection of known topics and their word distributions, Source-LDA is able to identify the subset of these topics that appear in a given corpus. 
It allows some variance in word distributions to the extent that it optimizes the topic modeling. 
\indent
A summary of the contributions of this work
are: \par
\begin{enumerate}
\item We propose a novel technique to topic modeling in a semi-supervised fashion that takes into account preexisting 
topic distributions.
\item We show how 
to find the appropriate topics in a corpus given an input set that contains a subset 
of the topics used to generate a corpus.
\item We explain how to make use of prior knowledge sources. In particular,
we show how to use Wikipedia articles to form word distributions.
\item We introduce an approach that allows for variance from an input topic to the latent topic discovered 
during the topic modeling process.
\end{enumerate}
\vspace{0mm} \par
\indent
The rest of this paper is organized as follows: In Section 2,
we give a brief introduction to the LDA algorithm and the Dirichlet distribution. A more detailed description of the Source-LDA
algorithm is presented in Section 3. In Section 4, the algorithm is
used and evaluated under various metrics.
Related literature is highlighted in Section 5. Section 6 gives the conclusions
of this paper. \par
\indent
\emph{For reproducible research, we make all of our code available online}.\footnote{\url{https://github.com/ucla-scai/Source-LDA}}
\begin{figure*}[!t]
\centering
\subfigure[LDA]  
{
  \label{fig:lda(a)}
  \begin{tikzpicture}
  \node[obs] (w) {$\boldsymbol{w}$};
  \node[latent, above=0.75cm of w] (z) {$\boldsymbol{z}$};
  \node[latent, above=0.75cm of z] (theta) {$\boldsymbol {\theta}$};
  \node[latent, right=1.5cm of theta] (alpha) {$\boldsymbol {\alpha}$};
  \node[latent, right=1.5cm of w] (phi) {$\boldsymbol {\phi}$};
  \node[latent, right=1.0cm of phi] (beta) {$\boldsymbol {\beta}$};
  
  \edge {z} {w} ;
  \edge {theta} {z} ;
  \edge {alpha} {theta} ;
  \edge {phi} {w} ;
  \edge {beta} {phi} ;
  
  \plate {} {(phi)} {$K$} ;
  \plate [xscale=0.05cm, opacity=0.0] {wdh} {(w)(z)} {} ;
  \plate {wd} {(w)(z)(wdh.north west)(wdh.north east)} {$N_d$} ;
  \plate [xscale=0.045cm, yscale=0.02cm, opacity=0.0] {dh} {(wd.north west)(wd.north east)(wd.south west)(wd.south east)} {} ;
  \plate {d} {(w)(theta)(z)(dh.north west)(dh.north east)(dh.south west)(dh.south east)} {$D$} ;
  \end{tikzpicture}
}
\hfill
\subfigure[Source-LDA]
{
  \begin{tikzpicture}

  \node[obs] (w) {$\boldsymbol{w}$};
  \node[latent, above=1.0cm of w] (z) {$\boldsymbol{z}$};
  \node[latent, above=1.0cm of z] (theta) {$\boldsymbol {\theta}$};
  \node[latent, right=1.9cm of theta] (alpha) {$\boldsymbol {\alpha}$};
  \node[latent, left=1.9cm of w] (phi_m) {$\boldsymbol {\phi_m}$};
  \node[latent, right=1.9cm of w] (phi_k) {$\boldsymbol {\phi_k}$};
  \node[latent, right=1.0cm of phi_k] (beta) {$\boldsymbol {\beta}$};
  \node[latent, above=1.0cm of phi_m] (lambda) {$\boldsymbol{\lambda}$};
  \node[obs,  left=0.7cm of lambda] (x) {$\boldsymbol{X}$};
  \node[obs,  left=1.0cm of x] (kb) {$\boldsymbol{KS}$};
  \node[latent,  below=1.0cm of x] (delta) {$\boldsymbol{\delta}$};
  \node[latent, above=1.0cm of lambda] (sigma) {$\boldsymbol {\sigma}$};
  \node[latent, above=0.9cm of x] (mu) {$\boldsymbol{\mu}$};

  \edge {z} {w} ;
  \edge {theta} {z} ;
  \edge {alpha} {theta} ;
  \edge {phi_m} {w} ;
  \edge {phi_k} {w} ;
  \edge {beta} {phi_k} ;
  \edge {kb} {x};
  \edge {x} {delta};
  \edge {mu} {lambda};
  \edge {sigma} {lambda};
  \edge {lambda} {delta};
  \edge {delta} {phi_m};

  \plate {} {(phi_m)(x)(lambda)} {$B$} ;
  \plate {} {(phi_k)} {$K$} ;
  \plate [xscale=0.065cm, opacity=0.0] {wdh} {(w)(z)} {} ;
  \plate {wd} {(w)(z)(wdh.north west)(wdh.north east)} {$N_d$} ;
  \plate [xscale=0.040cm, yscale=0.022cm, opacity=0.0] {dh} {(wd.north west)(wd.north east)(wd.south west)(wd.south east)} {} ;
  \plate {d} {(w)(theta)(z)(dh.north west)(dh.north east)(dh.south west)(dh.south east)} {$D$} ;
\end{tikzpicture}
}
\caption{Plate notation for LDA (a), and the proposed Source-LDA (b).}
\label{lda}
\end{figure*}
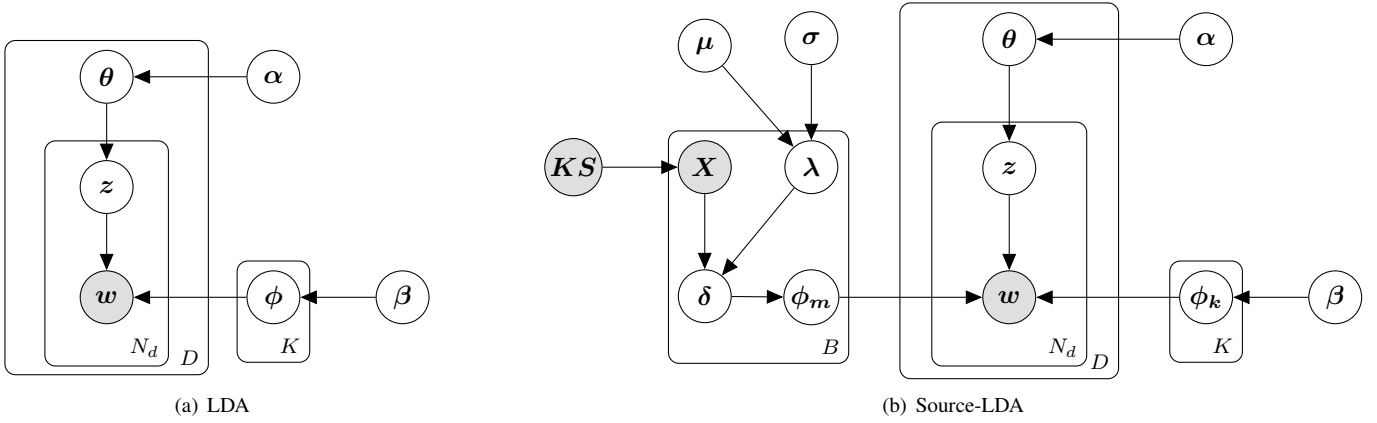
\section{Preliminaries}\label{mot}
\subsection{Dirichlet Distribution}
The Dirichlet distribution is a distribution over 
probability mass functions with a specific number of atoms
and is commonly used in Bayesian models. A property of
the Dirichlet that is often used in inference of Bayesian
models is conjugacy to the multinomial distribution. This
allows for the posterior of a random variable with a 
multinomial likelihood and a Dirichlet prior to also be
a Dirichlet distribution.\par
\indent The parameters are given as a vector denoted 
by $\alpha$. The probability density function for a given
probability mass function (PMF) $\theta$ and parameter vector $\alpha$ of
length $J$ is defined as:
\begin{displaymath}
f(\theta, \alpha) = \frac{\Gamma(\sum_i^J \alpha_i)}{\prod_i^J \Gamma(\alpha_i)}  \prod\limits_i^J \theta_i^{\alpha_i - 1} 
\end{displaymath}
\indent A sample from the Dirichlet distribution produces a PMF that is parameterized by $\alpha$. The choice
of a particular set of $\alpha$ values influences the outcome
of the generated PMF. If all $\alpha$ values are the same
(symmetric parameter), as $\alpha$ approaches $0$, the probability
will be concentrated on a smaller set of atoms. As $\alpha$ approaches
infinity, the PMF will become the uniform distribution.
If all $\alpha_i$ are natural numbers then each
individual $\alpha_i$ can be thought of as the ``virtual'' count
for the $i_{th}$ value~\cite{Minka00bayesianinference}.
\subsection{Latent Dirichlet Allocation}
Latent Dirichlet Allocation (LDA) is the basis for many existing probabilistic topic models, and the
framework for the approach presented by this paper. Since we enhance
the LDA model in our proposed approach it is worth giving a brief
overview of the algorithm and model of LDA.\par
\indent LDA is a hierarchical Bayes model which utilizes Dirichlet priors
to estimate the intractable latent variables of the model. At a high level,
LDA is based on a generative model in which each word of
an input document from a corpus is chosen by first selecting a topic
that corresponds to that word and then selecting the word from a
topic-to-word distribution. Each topic-to-word distribution and
word-to-topic distribution is drawn from its respective
Dirichlet distribution. The formal definition of the generative
algorithm over a corpus is:
\vspace{1.5mm}
\begin{spacing}{1.20}
\algrenewcommand{\alglinenumber}[1]{#1.\phantom{12}}
\begin{algorithmic}[1]
  \State For each of the $K$ topics $\phi_k$:
    \State\hspace{\algorithmicindent} Choose $\phi_k \sim \text{Dir}(\beta)$
  \State For each of the $D$ documents $d$:
    \State\hspace{\algorithmicindent} Choose $N_d \sim \text{Poisson}(\xi)$
    \State\hspace{\algorithmicindent} Choose $\theta_d \sim \text{Dir}(\alpha)$
    \State\hspace{\algorithmicindent} For each of the $N_d$ words $w_{n,d}$:
      \State\hspace{\algorithmicindent}\hspace{\algorithmicindent} Choose $z_{n,d} \sim \text{Multinomial}(\theta)$
      \State\hspace{\algorithmicindent}\hspace{\algorithmicindent} Choose $w_{n,d} \sim \text{Multinomial}(\phi_{z_{n,d}})$
\end{algorithmic}
\end{spacing}
\vspace{1.5mm}
From the generative algorithm the resultant Bayes model is
shown by Figure \ref{lda}(a).\par
\indent Bayes' law is used to infer the latent $\theta$ distribution,
$\phi$ distribution, and $z$
\vspace{1mm}
\begin{displaymath}
P(\theta,\phi,z|w,\alpha,\beta) = \frac{p(\theta,\phi,z,w|\alpha,\beta)}{p(w|\alpha,\beta)}
\end{displaymath}
\vspace{1mm}
Unfortunately the exact computation of this equation is intractable. Hence, it must be 
approximated with techniques such as expectation-maximization~\cite{Blei03latentdirichlet},
Gibbs sampling or collapsed Gibbs sampling~\cite{griffiths_steyvers04}.
\section{Proposed Approach}\label{app}
Source-LDA is an extension of the LDA generative model. In Source-LDA,
after a known set of topics are determined, an initial
word-to-topic distribution is generated from corresponding Wikipedia articles. 
The desiderata is to enhance existing LDA topic modeling
by integrating prior knowledge into the topic modeling process. 
The relevant terms and concepts used in the following discussion
are defined below. 
\begin{definition}[Knowledge source]
A knowledge source is a collection of documents that are focused on describing
a set of concepts. For example the knowledge source used in our
experiments are Wikipedia articles that
describe the categories we select
from the Reuters dataset.
\end{definition}
\begin{definition}[Source Distribution]
The source distribution is a discrete probability distribution
over the words of a document describing a topic. The probability mass function
is given by 
\[
  \forall w_i \in W\!\text{,\phantom{h}}f(w_i) = \frac{n_{w_i}}{\sum_j^G n_{w_j}}
\]
where $W$
is the set of all words in the document, $G = |W|$, and $n_{w_i}$ is the
number of times word $w_i$ appears in the document. 
\end{definition}
\begin{definition}[Source Hyperparameters]
For a given document in a knowledge source the
knowledge source hyperparameters are 
defined by the vector $(X_1, X_2, \ldots, X_V)$ where $X_i = n_{w_i} + \epsilon$
and $\epsilon$ is a very small positive number
that allows for non-zero probability draws from
the Dirichlet distribution.
$V$ is the size of the vocabulary of the corpus for which
we are topic modeling, and $n_{w_i}$ is the number of times the word $w_{i}$
from the corpus vocabulary appears in the knowledge source document.
\end{definition}
\par
\indent We detail three approaches to capture the
intent of Source-LDA. The first approach is a simple enhancement to the
LDA model that allows for the influencing of topic distributions, but 
suffers from needing more user intervention.
The second approach allows for the mixing of unknown topics, and
the third approach combines the previous two approaches. It
moves toward a complete solution to topic modeling based off
prior knowledge sources. 
\subsection{Bijective Mapping}
In the simplest approach, the Source-LDA model assumes that
there exists a 1-to-1 mapping between a known set of topics
and the topics used to generate a corpus.
The generative model then assumes that, instead of selecting
topic-to-word distributions from sampling from the Dirichlet distribution,
a set of $K$ distributions are given as input and sampled from after each 
topic assignment is sampled for a given token position.
The generative process for a corpus adapted from the traditional LDA 
generative model during the construction of the $\phi$ 
distributions is as follows (for brevity only the relevant parts of 
the existing LDA algorithm are shown):
\vspace{2mm}
\begin{spacing}{1.20}
\algrenewcommand{\alglinenumber}[1]{#1.\phantom{12}}
\begin{algorithmic}[1]
  \State For each of the $K$ topics $\phi_k$:
    \State\hspace{\algorithmicindent} $\delta_k \gets (X_{k,1}, X_{k,2}, \ldots, X_{k,V})$
    \State\hspace{\algorithmicindent} Choose $\phi_k \sim \text{Dir}(\delta_k)$
\end{algorithmic}
\end{spacing}
\vspace{1mm}
\noindent Where $(X_{k,1}, X_{k,2}, \ldots, X_{k,V})$ represents the
knowledge source hyperparameters for the $k_{\text{th}}$ knowledge source document.
The generative model only differs from the traditional
LDA model in how each $\phi$ is built. Therefore the
derivation for inference is a simple factor as well.
To approximate the distributions for $\theta$ and $\phi$,
a collapsed Gibbs sampler can approximate the $z$
assignments as follows:
\begin{displaymath}
P(w_i|z_i\text{=}j,z_{\text{-}i},w_{i}) \propto P(w_i | z_i\text{=}j, z_{\text{-}i},w_{\text{-}i})P(z_i\text{=}j|z_{\text{-}i})
\end{displaymath}
\noindent From the Bayesian Model the following equations can be
easily be generated
\begin{displaymath}
P(w_i|z_i\text{=}j,z_{\text{-}i},w_{\text{-}i})\!=\!\!\int\!\!P(w_i | z_i\text{=}j, \phi_j)P(\phi_j|z_{\text{-}i},w_{\text{-}i})d\phi_j
\end{displaymath}
\noindent with
\begin{displaymath}
P(\phi_j|z_{\text{-}i},w_{\text{-}i}) \propto P(w_{\text{-}i} | \phi_j, z_{\text{-}i})P(\phi_j)
\end{displaymath}
\begin{displaymath}
P(\phi_j|z_{\text{-}i},w_{\text{-}i}) = Dir(\delta_{i,j} + n_{w_{\text{-}i,j}})
\end{displaymath}
\begin{displaymath}
P(w_i | z_i\text{=}j, \phi_j) = \phi_{w_{i,j}}
\end{displaymath}
\begin{displaymath}
P(w_i|z_i\text{=}j,z_{\text{-}i},w_{\text{-}i})\!=\!Dir(\delta_{i,j} + n_{w_{\text{-}i,j}})\!\int\!\!\phi_{w_{i,j}}d\phi_j
\end{displaymath}
\begin{displaymath}
P(w_i|z_i\text{=}j,z_{\text{-}i},w_{\text{-}i})\!=\!\frac{n_{\text{-}i,j}^{w_i} + \delta_{i,j}}{n_{\text{-}i,j}^{(\cdot)} + \sum_a^V\delta_{a,j}}
\end{displaymath}
$n^{w}$ and $n^{d}$ in this and the following equations represent a count matrix
for the number of times a word is assigned to
a topic and the number of times a topic is assigned
to a document respectively. For brevity since the prior probability is unchanged
in the ``Bijective Mapping'' model we will skip the derivation
which is well defined in other articles~\cite{griffiths_steyvers04, darling2011theoretical, griffiths2002gibbs}.
\begin{displaymath}
P(z_i\text{=}j|z_{\text{-}i})\!=\!\frac{n_{\text{-}i,j}^{d_i} + \alpha}{n_{\text{-}i}^{(d_i)} + K\alpha}
\end{displaymath}
Putting the two equations together gives the final
Gibbs sampling equation:
\begin{displaymath}
P(z_i\text{=}j|z_{\text{-}i},w) \propto \frac{n_{\text{-}i,j}^{w_i} + \delta_{i,j}}{n_{\text{-}i,j}^{(\cdot)} + \sum_a^V\delta_{a,j}} \frac{n_{\text{-}i,j}^{d_i} + \alpha}{n_{\text{-}i}^{(d_i)} + K\alpha}
\end{displaymath}
\noindent Given the approximation to the topic assignments,
the $\theta$ and $\phi$ distributions are calculated as:
\begin{displaymath}
\phi_{w,t} = \frac{n_{w,t} + \delta_{w,t}}{n_{t} + \sum_a^V\delta_{a,t}}
\end{displaymath}
\begin{equation}
\theta_{t,d} = \frac{n_{d,t} + \alpha}{n_{d} + K\alpha}\label{eq:theta}
\end{equation}
\begin{figure}[!t]
\centering
\includegraphics[trim={0.25cm 0.20cm 0.21cm 0.19cm},clip,width=70mm]{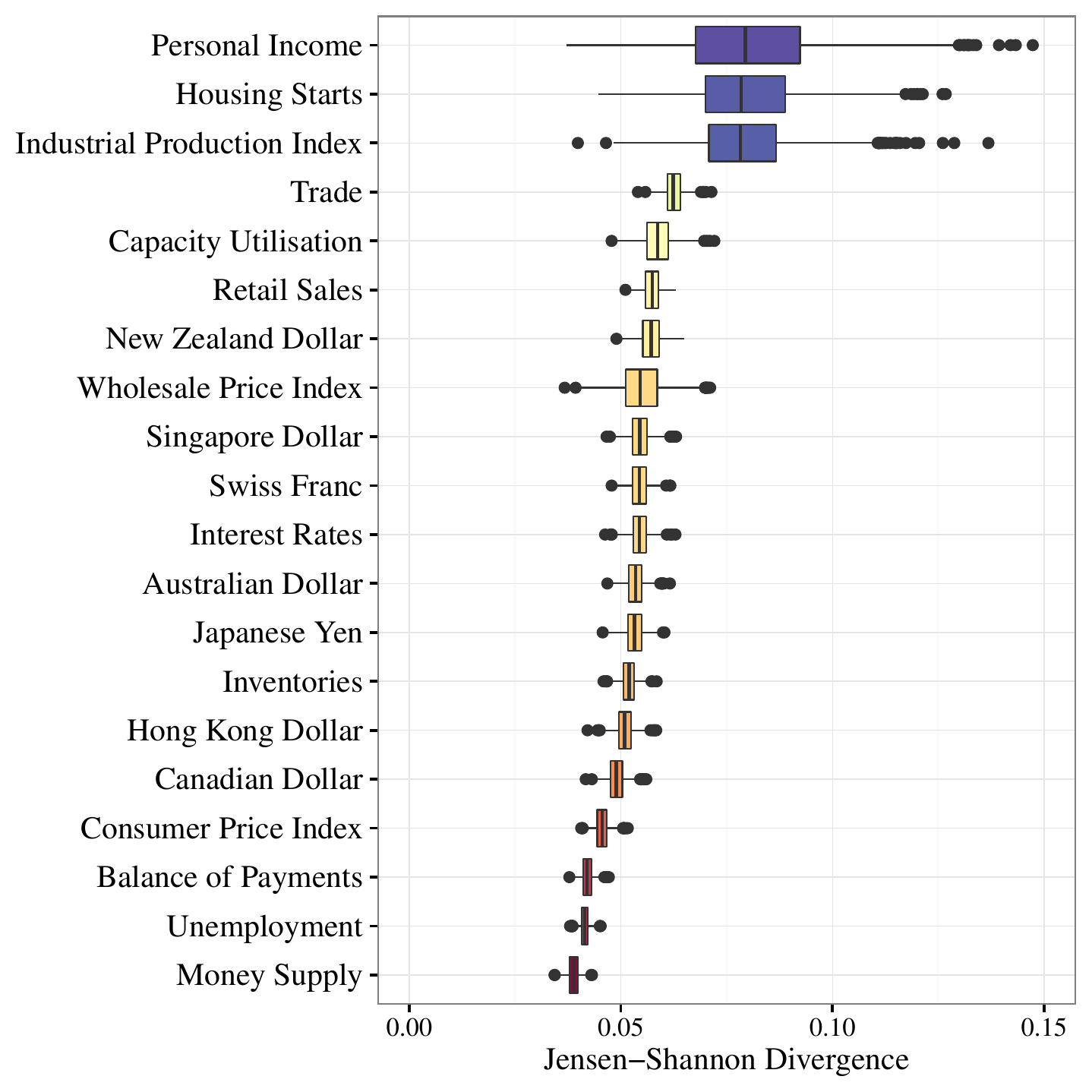}
\caption{Box plots~\cite{github} showing the Jensen\text{-}Shannon divergence 
(the JS divergence measures the distance or similarity 
between probability distributions) of 
1000 Dirichlet samples parameterized by source 
hyperparameters for a subset of knowledge source topics. The topics
were taken from Wikipedia pages.}
\label{fig:thousand}
\end{figure} \par
\indent
In the case when all topics are known, this model has
the advantage of conforming the $\phi$ distributions to 
the source distributions, but has three drawbacks. First,
even though there is some variability between the
$\phi$ distribution and source distribution, as illustrated
by Figure \ref{fig:thousand}, there may be cases in which this constraint 
should be relaxed even further. This is because it is
entirely possible to generate a corpus about a known topic
without exactly following the frequencies at which the 
topic is discussed in its respective article.
This model also requires the user to input
the known topics, and other possible supervised
approaches may be better suited to the task~\cite{slda, disclda, labeledlda}. The third drawback is
that we are not allowing the possibility that the corpus
was generated from a mixture of known topics and unknown topics,
which is a more realistic scenario for an arbitrary document. 
The next model aims to resolve this last deficiency.\par
\subsection{Known Mixture of Topics}
The next model assumes that in the topic model it
is given how many topics are known topics
(as well as their word distributions)
and how many are unknown topics.
The previous approach works quite well in this situation
in that an unknown topic will have a symmetric beta
parameter which will capture assignments which were
unallocated due to a low probability in matching any known topic.\par
\indent The resulting model helps to solve the existing problems
of the bijective model and only requires a minor input
to the existing generative model. The resulting model
works quite well with the bijective model in that the
symmetric Dirichlet prior can be used to guide a topic
toward being a general unknown topic or a known topic.
The model changes as shown below with a minor change to
the generative algorithm and the collapsed Gibbs sampling.
\vspace{1mm}
\begin{spacing}{1.20}
\algrenewcommand{\alglinenumber}[1]{#1.\phantom{12}}
\begin{algorithmic}[1]
  \State For each of the $K$ topics $\phi_k$:
    \State\hspace{\algorithmicindent} \textbf{if} $k \leq T$ \textbf{then}  
      \State\hspace{\algorithmicindent}\hspace{\algorithmicindent} Choose $\phi_k \sim \text{Dir}(\beta)$
    \State\hspace{\algorithmicindent} \textbf{else}
    \State\hspace{\algorithmicindent}\hspace{\algorithmicindent} $\delta_k \gets (X_{k,1}, X_{k,2}, \ldots, X_{k,V})$
      \State\hspace{\algorithmicindent}\hspace{\algorithmicindent} Choose $\phi_k \sim \text{Dir}(\delta_k)$
\end{algorithmic}
\end{spacing}
\vspace{1mm}
\noindent Where $T$ is the total number of 
non-source topics. The change required to the collapsed Gibbs 
sampling is then:
\begin{displaymath}
P(z_i\text{=}j|z_{\text{-}i},w) \propto \frac{n_{\text{-}i,j}^{w_i} + \beta}{n_{\text{-}i,j}^{(\cdot)} + W\beta}
\frac{n_{\text{-}i,j}^{d_i} + \alpha}{n_{\text{-}i}^{(d_i)} + K\alpha} \text{,\phantom{h}} \forall i \leq T\label{eq:gibbs}
\end{displaymath}
\noindent and
\begin{equation}
P(z_i\text{=}j|z_{\text{-}i},w) \propto \frac{n_{\text{-}i,j}^{w_i} + \delta_{i,j}}{n_{\text{-}i,j}^{(\cdot)} + \sum_a^V\delta_{a,j}}
\frac{n_{\text{-}i,j}^{d_i} + \alpha}{n_{\text{-}i}^{(d_i)} + K\alpha} \text{,\phantom{h}} \forall i > T
\end{equation}
\noindent This approach gives the benefit of allowing a 
mixture of known topics and unknown topics, but problems
still arise in that the Dirichlet distributions for the
source distribution may be too restricting.
\subsection{Source-LDA}
By using the counts as hyperparameters,
the resultant $\phi$ distribution will take on the shape
of the word distribution derived from the knowledge source. However, this might be at odds with the
aim of enhancing existing topic modeling. With the
goal to influence the $\phi$ distribution, it is entirely plausible
to have divergence between the two distributions.
In other words, $\phi$ may not need to strictly follow the
corresponding knowledge source distribution. 
\subsubsection{Variance from the source distribution} To allow for this relaxation,
another parameter $\lambda$ is introduced into the model which
is used to allow for a higher deviance from the source
distribution. To obtain this variance each source hyperparameter will be
raised to a power of $\lambda$. Thus as $\lambda$ approaches $0$ each hyperparameter will approach $1$ and the
subsequent Dirichlet draw will allow all discrete distributions with equal probability.
As $\lambda$ approaches $1$
the Dirichlet draw will be tightly
conformed to the source distribution. \par
\indent The addition of $\lambda$ changes the existing
generative model only slightly
and allows for a variance for each individual
$\delta_i$, which frees us from an overly restrictive
binding to the associated knowledge source distribution. 
The $\lambda$ parameter acts as a measure of how much divergence
is allowed for a given modeled topic from the knowledge source
distribution. Figure \ref{fig:single_nonsmooth} shows how the JS Divergence changes
with changes to the $\lambda$ parameter.\par
\vspace{2mm}
\begin{spacing}{1.25}
\algrenewcommand{\alglinenumber}[1]{\the\numexpr #1+4 .\phantom{12}}
\begin{algorithmic}[1]
    \State\hspace{\algorithmicindent}\hspace{\algorithmicindent} $\delta_k \gets [(X_{k,1})^\lambda, (X_{k,2})^\lambda, \ldots, (X_{k,V})^\lambda]$
\end{algorithmic}
\end{spacing}
\indent With the introduction of $\lambda$ as an input
parameter, the new topic model has the advantage of
allowing variance and also leaves the collapsed 
Gibbs sampling equation unchanged. However this
also requires a uniform variance from the knowledge base distribution
for all latent topics. This can be a problem if the corpus
was generated with some topics influenced strongly while others
less so. To solve this we can introduce $\lambda$ as a hidden
parameter of the model. \par
\subsubsection {Approximating $\lambda$}
In the ideal situation $\lambda$ will be as close to $1$
for most knowledge based latent topics, with the flexibility to
deviate as required by the data. For this we assume a 
Gaussian prior over $\lambda$ with mean set to $\mu$. The
variance then becomes a modeled parameter that
conceptually can be thought of as how much variance from
the knowledge source distribution we wish to allow in our topic model. In
assuming a Gaussian prior for $\lambda$, we must
integrate $\lambda$ out of the collapsed Gibbs sampling
equations (only the probability of $w_i$ under 
topic $j$ is shown, the probability of topic $j$ in 
document $d$ is unchanged and omitted).
\begin{figure}[!t]
\centering
\includegraphics[trim={0.10cm 0.85cm 1.0cm 2.0cm},clip,width=70mm]{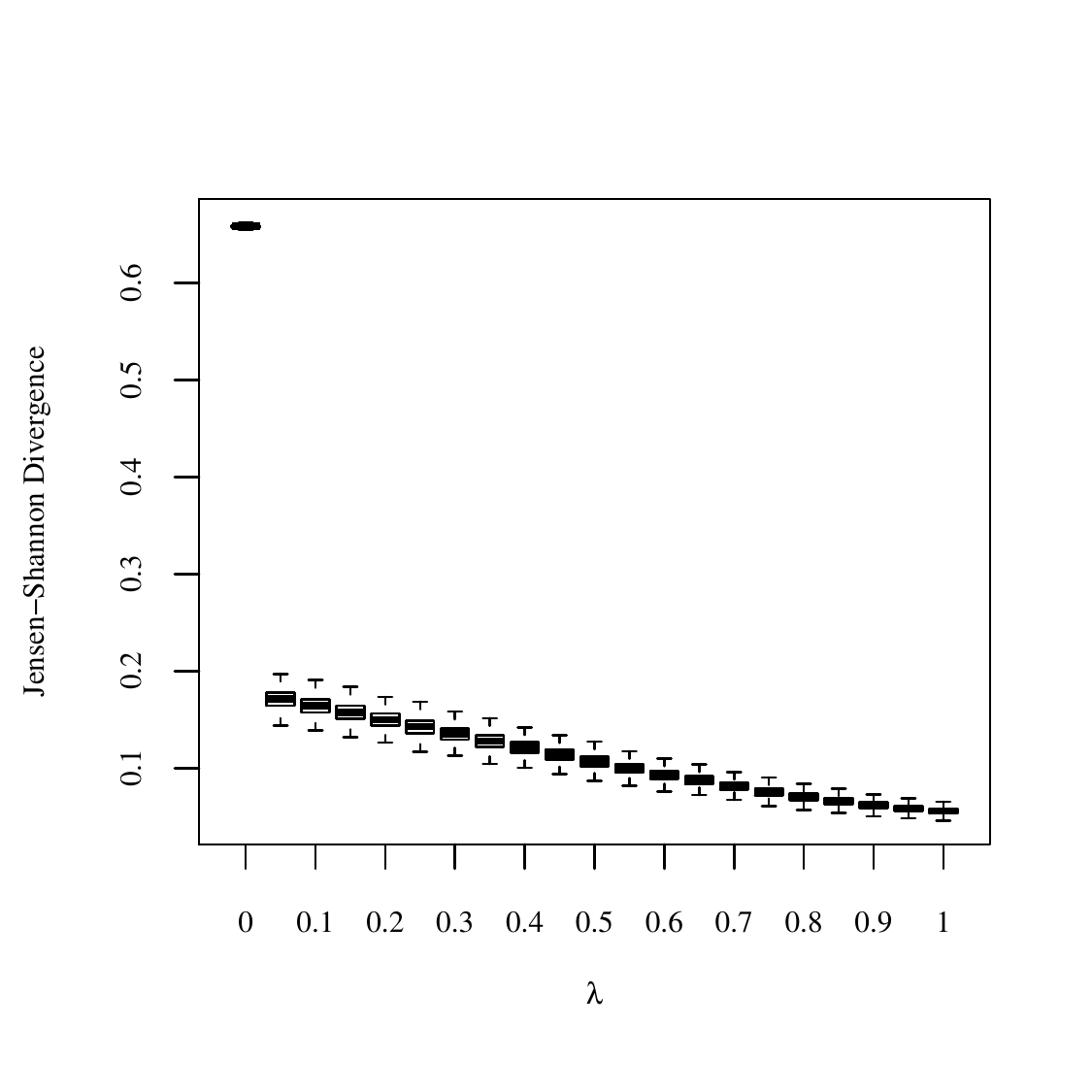}
\caption{Box plots showing how the JS divergence between a source distribution and a Dirichlet sample parameterized by
source hyperparameters raised to $\lambda$ changes with changes to $\lambda$ without smoothing.}
\label{fig:single_nonsmooth}
\end{figure}
\FloatBarrier
\begin{figure}[!t]
\centering
\includegraphics[trim={0.10cm 0.85cm 1.0cm 2.0cm},clip,width=70mm]{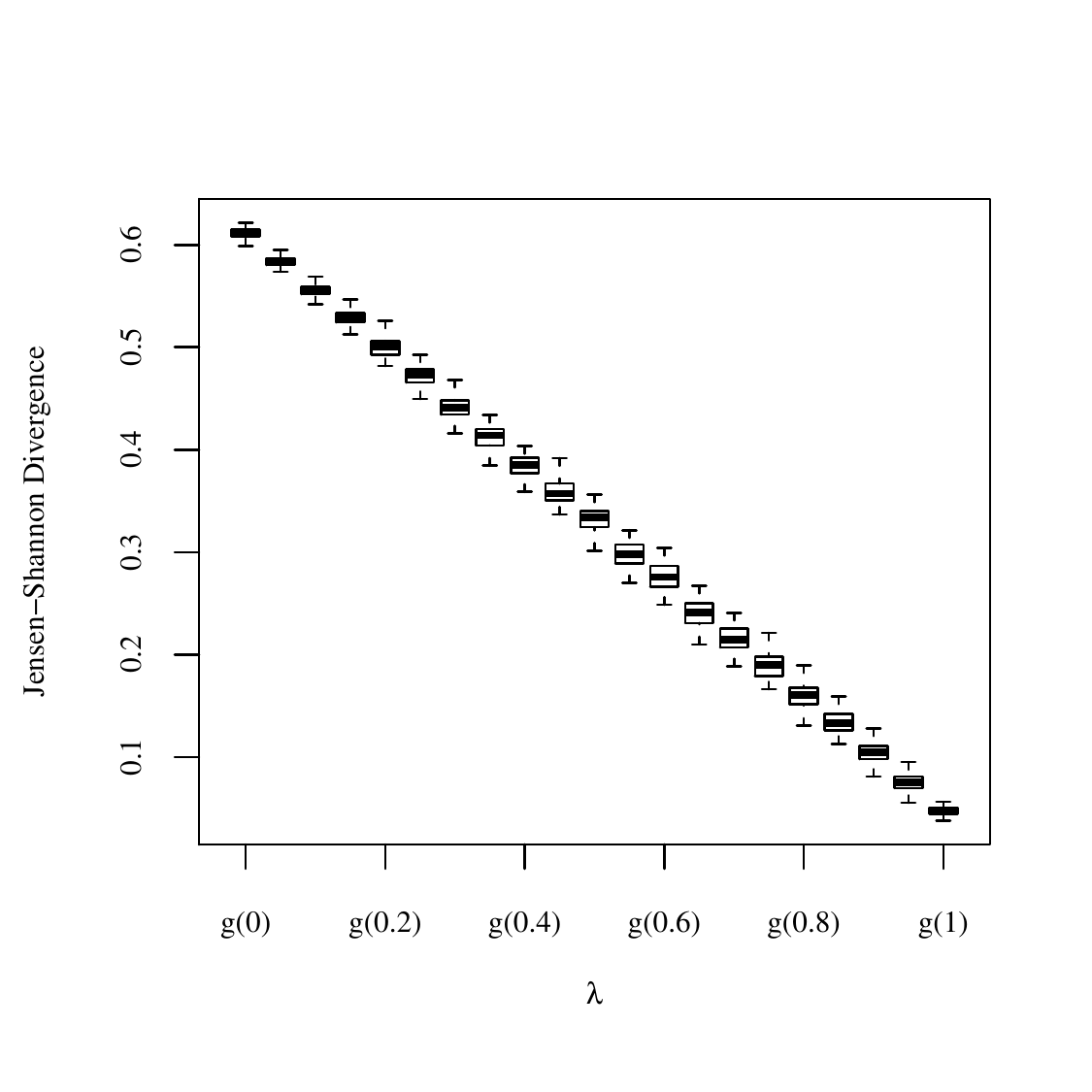}
\caption{The JS divergence between a source distribution and a Dirichlet sample parameterized by
source hyperparameters raised to $\lambda$ with $\lambda$
mapped to a linear smoothing function $g$.}
\label{fig:single_smoothed}
\end{figure}
\begin{displaymath}
P(z_i\text{=}j|z_{\text{-}i},w) \propto \int \frac{n_{\text{-}i,j}^{w_i} + (\delta_{i,j})^\lambda}{n_{\text{-}i,j}^{(\cdot)} + \sum_a^V(\delta_{a,j})^\lambda} \mathcal{N}(\mu, \sigma) d\lambda
\end{displaymath}
\noindent $\phi$ then becomes
\begin{displaymath}
\phi_{w,t} = \int \frac{n_{w,t} + (\delta_{w,t})^\lambda}{n_{t} + \sum_a^V(\delta_{a,t})^\lambda} \mathcal{N}(\mu, \sigma) d\lambda
\end{displaymath}
\noindent Unfortunately closed form expressions for these integrals are hard to obtain 
and so they must be approximated numerically during sampling. \par
\indent Another problem arises in that the change of $\lambda$ is
not in par with the change of the Gaussian distribution, as can
be seen in Figure \ref{fig:single_nonsmooth}.
To make the changes of $\lambda$ more in line
with that expected from the Gaussian PDF, we must map each
individual $\lambda$ value
in the range $0$ to $1$ with a value which produces
a change in the JS divergence in a linear fashion. 
We approximate a function, $g(x)$ with a linear derivative, shown
in Figure \ref{fig:single_smoothed}.
The approach taken to approximate $g(x)$ is by linear
interpolation of an aggregated large number of samples for each point taken
in the range $0$ to $1$.
Our collapsed Gibbs sampling equations then becomes:
\begin{equation}
P(z_i\text{=}j|z_{\text{-}i},w) \propto \int \frac{n_{\text{-}i,j}^{w_i} + (\delta_{i,j})^{g(\lambda)}}{n_{\text{-}i,j}^{(\cdot)} + \sum_a^V(\delta_{a,j})^{g(\lambda)}} \mathcal{N}(\mu, \sigma) d\lambda
\end{equation}
\begin{displaymath}
\phi_{w,t} = \frac{n_{w,t} + \beta}{n_{t} + V\beta} \text{,\phantom{h}} \forall t \leq T
\end{displaymath}
\noindent and
\begin{equation}\label{eq:phi}
\phi_{w,t} = \int \frac{n_{w,t} + (\delta_{w,t})^{g(\lambda)}}{n_{t} + \sum_a^V(\delta_{a,t})^{g(\lambda)}} \mathcal{N}(\mu, \sigma) d\lambda \text{,\phantom{h}} \forall t > T
\end{equation}
%
%
%
\subsubsection{Superset Topic Reduction} A third problem 
involves knowing the right mixture
of known topics and unknown topics. It is also entirely 
possible that many known topics may not be used by the
generative model. Our desire to leave the model as 
unsupervised as possible calls for input that is a 
superset of the actual generative topic selection 
in order to avoid manual topic selection. 
In the case of modeling only a specific
number of topics over the corpus, the problem then becomes how to
choose which knowledge source latent topics to allow in the model vs. how many unlabeled
topics to allow.\par
\indent The goal then is to allow for a
superset of knowledge source topics as input and then during the 
inference to select the best subset of these with a mixture
of unknown topics where the total number of unlabeled topics is
given as input $K$. The approach given is to
use a mixture of $K$ unlabeled topics alongside
the labeled knowledge source topics. The total number of topics then becomes $T$.
During the inference we eliminate topics which are not assigned
to any documents. At the end of the sampling phase we
then can use a clustering algorithm (such as k-means, JS divergence) to further reduce the modeled
topics and give a total of $K$ topics. As described more in the experimental section, with the goal of
capturing topics that were frequently occurring in the corpus, topics not appearing in a frequent 
enough of documents were eliminated. \par

\indent The complete generative process is shown in Figure \ref{lda}(b) and described below:
\vspace{1.5mm}
\begin{spacing}{1.20}
\algrenewcommand{\alglinenumber}[1]{#1.\phantom{12}}
\begin{algorithmic}[1]
  \State For each of the $T$ topics $\phi_t$:
    \State\hspace{\algorithmicindent} \textbf{if} $t \leq K$ \textbf{then}  
      \State\hspace{\algorithmicindent}\hspace{\algorithmicindent} Choose $\phi_t \sim \text{Dir}(\beta)$
    \State\hspace{\algorithmicindent} \textbf{else}
      \State\hspace{\algorithmicindent}\hspace{\algorithmicindent} Choose $\lambda_t \sim \mathcal{N}(\mu, \sigma)$
      \State\hspace{\algorithmicindent}\hspace{\algorithmicindent} $\delta_t \gets [(X_{t,1})^{g(\lambda_t)}, (X_{t,2})^{g(\lambda_t)}, \ldots, (X_{t,V})^{g(\lambda_t)}]$
      \State\hspace{\algorithmicindent}\hspace{\algorithmicindent} Choose $\phi_t \sim \text{Dir}(\delta_t)$
  \State For each of the $D$ documents $d$:
    \State\hspace{\algorithmicindent} Choose $N_d \sim \text{Poisson}(\xi)$
    \State\hspace{\algorithmicindent} Choose $\theta_d \sim \text{Dir}(\alpha)$
    \State\hspace{\algorithmicindent} For each of the $N_d$ words $w_{n,d}$:
      \State\hspace{\algorithmicindent}\hspace{\algorithmicindent} Choose $z_{n,d} \sim \text{Multinomial}(\theta)$
      \State\hspace{\algorithmicindent}\hspace{\algorithmicindent} Choose $w_{n,d} \sim \text{Multinomial}(\phi_{z_{n,d}})$
\end{algorithmic}
\end{spacing}
\vspace{1.5mm}
The full collapsed Gibbs sampling algorithm is given in algorithm \ref{alg:c_gibbs}.
\subsubsection {Analysis}
By using a clustering algorithm or thresholding the topic document frequency,
the collapsed Gibbs algorithm is guaranteed to produce $K$ topics. The running time is a function
of the number of iterations $I$, average words per document $D_{\text{avg}}$, 
number of documents $D$, number of topics $T$ and number of
approximation steps $A$, and
is $\mathcal{O}(I \times D_{\text{avg}} \times D \times T \times A)$.
This differs only from the traditional collapsed Gibbs sampling
in LDA by an increase of $(T-K)A$. But since we have built the
approach to potentially have a large $T-K$ this difference can have a
significant impact on running times. \par

\indent Approaches exist that
can parallelize the sampling procedure, but these are
often approximations or can potentially have slower than baseline
running times~\cite{plda, adlda, fastlda}. We present two modifications to the original
algorithm that allow for inference while guaranteeing
the exactness of the results to the original Gibbs sampling.
The first one makes use of prefix sums rules~\cite{Blelloch90prefixsums} and guarantees 
a running time of:
\begin{displaymath}
\mathcal{O}(I \times D_{\text{avg}} \times D \times A \times Max[T/P,P])
\end{displaymath}
with $P$ being
the number of parallel units. This algorithm is given by Algorithm \ref{alg:prefix_sum}.

\algdef{SE}[DOWHILE]{Do}{doWhile}{\algorithmicdo}[1]{\algorithmicwhile\ #1}
\begin{algorithm}[!t]
\scriptsize
\caption{Collapsed Gibbs}
\label{alg:c_gibbs}
{\bf Input:} Dirichlet hyperparameters $\alpha$, $\beta$, a corpus $C$, vocabulary $V$, unlabeled topic count $K$, total topic count $T$, a set of source topics $S$, mean $\mu$, variance $\sigma$, and iteration count $I$.\\
{\bf Output:} $\theta$, $\phi$\\
\begin{algorithmic}
  \Procedure{Collapsed\_Gibbs}{$\alpha$, $\beta$, $C$, $V$, $T$}
  \For {$t=K+1$ to $T$}\
    \State Calculate $g_t$
  \EndFor
  \State Initialize $C_\text{topics}$ to random topic assignments
  \State Update $n^w$ and $n^d$ from $C_\text{topics}$
  \For {$iter=1$ to $I$}\
    \For {$i=1$ to $C$}\
      \For {$j=1$ to $|C_i|$}\
        \State $C_{\text{topics}_{i,j}} \gets Sample(i,j)$
      \EndFor
    \EndFor
  \EndFor
  \State Calculate $\theta$ according to \textbf{Equation 1}
  \State Calculate $\phi$ according to \textbf{Equation 4}
  \State \Return $\theta, \phi$
  \EndProcedure \\
  \Procedure{Sample}{$i$, $j$}
  \State Decrement $n^w$ and $n^d$ accordingly
  \For {$t=1$ to $K$}\
    \State Calculate $p_t$ according to \textbf{Equation 2}
  \EndFor
  \For {$t=K+1$ to $T$}\
    \State Calculate $p_t$ according to \textbf{Equation 3}
  \EndFor
  \State $topic \sim \text{Multinomial}(p)$
  \State Increment $n^w$ and $n^d$ accordingly
  \State \Return $topic$
  \EndProcedure

\end{algorithmic}
\end{algorithm}
\algdef{SE}[DOWHILE]{Do}{doWhile}{\algorithmicdo}[1]{\algorithmicwhile\ #1}
\begin{algorithm}[!t]
\scriptsize
\caption{Prefix Sums Parallel Sampling}
\label{alg:prefix_sum}
\begin{algorithmic}
  \Procedure{Sample}{$i$, $j$}
    \State Decrement $n^w$ and $n^d$ accordingly
    \For {$i$ from $0$ to $T - 1$ in parallel}\
      \If {$i \leq K$}
        \State Calculate $p_i$ according to \textbf{Equation 2}        
      \Else
        \State Calculate $p_i$ according to \textbf{Equation 3}
      \EndIf
      \State $p_i\gets p_{i-1} + p_i$ 
    \EndFor
    \For {$d$ from $0$ to $(\ln T) - 1$}\
      \For {$i$ from $0$ to $T - 1$ by $2^{d+1}$ in parallel}\
        \State $p_{(i + 2^{d+1} - 1)} \gets p_{(i + 2^d - 1)} + p_{(i + 2^{d+1} - 1)}$ 
      \EndFor  
    \EndFor
    \State $p_{(T - 1)} \gets 0$ 
    \For {$d$ from $(\ln T) - 1$ down to $0$}\
      \For {$i$ from $0$ to $T - 1$ by $2^{d+1}$ in parallel}\
        \State $h \gets p_{(i + 2^d - 1)}$ 
        \State $p_{(i + 2^{d+1} - 1)} \gets p_{(i + 2^{d+1} - 1)}$ 
        \State $p_{(i + 2^{d+1} - 1)} \gets h + p_{(i + 2^{d+1} - 1)}$ 
      \EndFor  
    \EndFor
    \State $topic \gets \text{Binary\_Search}(p)$
    \State Increment $n^w$ and $n^d$ accordingly
    \State \Return $topic$
  \EndProcedure
\end{algorithmic}
\end{algorithm}
This algorithm is practical in situations where $T-K$ is
large, but suffers from the limitations of the number of
context switches required for the threads to wait at their
respective barriers. A simpler implementation approach that
reduces the number of context switches is to add
the sums for each thread then wait for a barrier. When the barrier
is released we add the end values together and 
then in parallel we add the remaining necessary items. This approach is
given in Algorithm \ref{alg:simple_parallel}. The running time is then:
\begin{displaymath}
\mathcal{O}(I \times D_{\text{avg}} \times D \times A \times Max[T/P,P])
\end{displaymath}
\par
\indent These two algorithms allow for mitigation of the increase in the number of
topics and should approach times very similar to those of standard
LDA runs. They are also very extensible and can be used in other optimization algorithms.

\algdef{SE}[DOWHILE]{Do}{doWhile}{\algorithmicdo}[1]{\algorithmicwhile\ #1}
\begin{algorithm}[!t]
\scriptsize
\caption{Simple Parallel Sampling}
\label{alg:simple_parallel}
\begin{algorithmic}
  \Procedure{Sample}{$i$, $j$}
    \State Decrement $n^w$ and $n^d$ accordingly
    \For {$i$ from $0$ to $T - 1$ in parallel}\
      \If {$i \leq K$}
        \State Calculate $p_i$ according to \textbf{Equation 2}        
      \Else
        \State Calculate $p_i$ according to \textbf{Equation 3}
      \EndIf
      \State $p_i\gets p_{i-1} + p_i$ 
    \EndFor
    \For {$i$ from $0$ to $T - 1$ by $T/P$}\
      \State $p_i\gets p_{(i-T/P)} + p_i$ 
      \State $ends_i\gets p_i$ 
    \EndFor

    \For {$i$ from $0$ to $T - 1$ in parallel}\
      \State $diff \gets p_{end} - ends_i$ 
      \State $p_i\gets diff + p_i$ 
    \EndFor

    \State $topic \gets \text{Binary\_Search}(p)$
    \State Increment $n^w$ and $n^d$ accordingly
    \State \Return $topic$
  \EndProcedure
\end{algorithmic}
\end{algorithm}

\subsubsection {Input determination}
Determining the necessary parameters and inputs
into LDA is an established research
area~\cite{wallach}, but since the
proposed model introduces additional input
requirements a brief overview will be given about
how to best set the parameters and determine the
knowledge source.
\paragraph{Parameter selection}
To determine the appropriate parameters, techniques
utilizing log likelihood have previously
been established~\cite{griffiths_steyvers04}. Since
these approaches generally require held out data
and are a function of the $\phi$, $\theta$, and $\alpha$
variables the introduction of $\lambda$ and $\sigma$
will not differentiate from their
original equations.
For example the perplexity calculations used for
Source-LDA are based off of importance
sampling~\cite{imp_2}, or latent variable
estimation via Gibbs sampling~\cite{hein}. Importance
sampling is only a function of $\phi$ given by Equation \ref{eq:phi},
and estimation via Gibbs sampling can made
using Equation \ref{eq:phi} and by the
following equation ($\tilde{z}$, $\tilde{w}$,
and $\tilde{n}$ represent the corresponding
variables in the test document set):
\begin{displaymath}
P(\tilde{z_i}\text{=}j|\tilde{z}_{\text{-}i},\tilde{w}) \propto \frac{n_{j}^{w_i} + \tilde{n}_{\text{-}i,j}^{w_i} + \beta}{n_{j}^{(\cdot)} + \tilde{n}_{\text{-}i,j}^{(\cdot)} + W\beta}
\frac{\tilde{n}_{\text{-}i,j}^{d_i} + \alpha}{\tilde{n}_{\text{-}i}^{(d_i)} + K\alpha} \text{,\phantom{h}} \forall i \leq T
\end{displaymath}
\noindent and
\begin{displaymath}
P(\tilde{z_i}\text{=}j|\tilde{z}_{\text{-}i},\tilde{w}) \propto \frac{n_{j}^{w_i}\! + \tilde{n}_{\text{-}i,j}^{w_i} + \delta_{i,j}}{n_{j}^{(\cdot)}\! + \tilde{n}_{\text{-}i,j}^{(\cdot)} + \sum\limits_a^V\delta_{a,j}}
\frac{\tilde{n}_{{\text{-}}i,j}^{d_i}\!+\alpha}{\tilde{n}_{\text{-}i}^{(d_i)}\!+\!K\alpha} \text{,}\thinspace \forall i > T
\end{displaymath}\par
\indent It is recommended to set the parameters so as to maximize
the log likelihood. Further analysis such as whether
or not the parameters can be learned a priori from
the data
are not the focus of this paper and are thus left as
an open research area.
\paragraph{Knowledge source selection}
Source-LDA is designed to be used only with
a corpus which has a known super set of topics which
comprise a large portion of the tokens. An example of such
a case is that of a corpus consisting of clinical patient
notes. Since there are extensive knowledge sources
comprising essentially all medical topics, Source-LDA
can be useful in discovering and labeling these existing
topics. In cases where it is not so easy to collect
a superset of topics traditional approaches may be more useful.
\section{Evaluation}\label{eva}
To test the results of the Source-LDA algorithm
we set up experiments to test against competing models. 
The most similar models to our proposed approach were used in comparison.
These are: latent Dirichlet allocation (LDA)~\cite{Blei03latentdirichlet}, explicit Dirichlet allocation (EDA)~\cite{explicit}, 
and the Concept-topic model (CTM)~\cite{concept_topic_modeling}.
Other approaches such as supervised latent Dirichlet allocation (sLDA)~\cite{slda}, discriminative LDA (DiscLDA)~\cite{disclda},
and labeled LDA (L-LDA)~\cite{labeledlda} are not used since a main desiderata of Source-LDA is
to require much less supervision than what is needed by these methods.
Likewise hierarchical methods~\cite{not_used} are omitted because
there is no established hierarchy in the knowledge source data for this model.
We describe in more detail below the experimental setups and
metrics used to compare results.
\begin{figure}[!t]
\centering
\subfigure[]
{
   \includegraphics[width=1\linewidth]{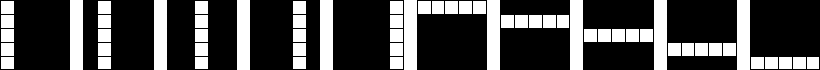}   
}
\hfill
\subfigure[]
{
   \includegraphics[width=1\linewidth]{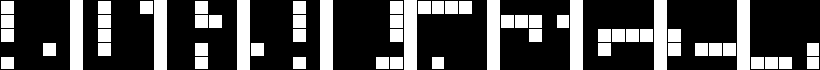}
}
\caption{A graphical representation of topics
containing 1 word for the cell locations of row and
column vectors in a 5 x 5 picture (a) and their augmented topics
after swapping a random assigned word (pixel) with a random topic's assigned word (b).}
\label{fig:graph_topics}
\end{figure}
\begin{figure}[!t]
\centering
\includegraphics[trim={0.0cm 0.1cm 0.0cm 0.35cm},clip,width=1\linewidth]{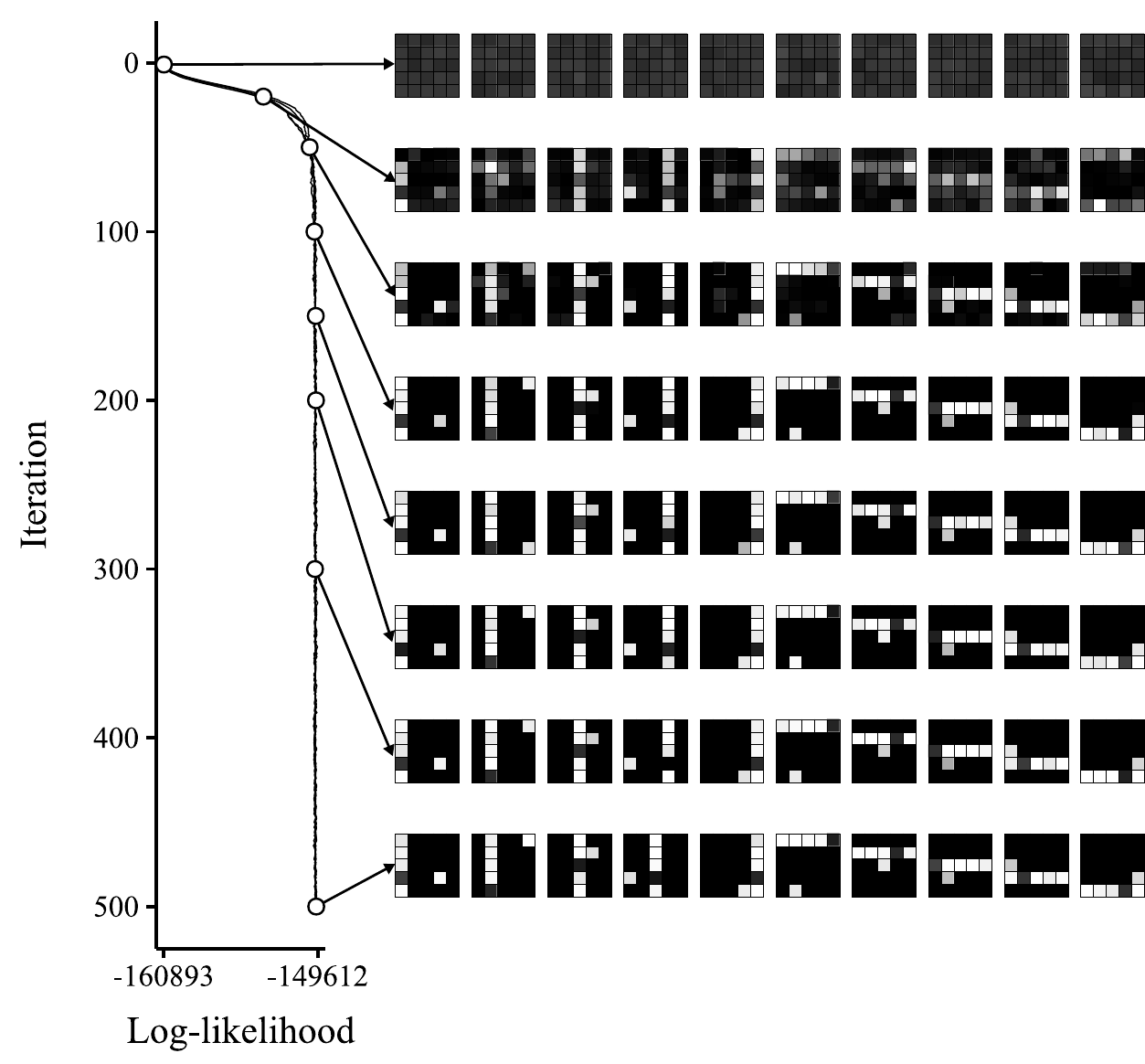}
\caption{Results from running Source-LDA for a 
corpus generated from topics in
Figure \ref{fig:graph_topics}(b)
using a knowledge source of topics
corresponding to
Figure \ref{fig:graph_topics}(a). Four separate
runs are plotted to show the similarity of the log-likelihood
relation to the iteration between the runs. The topics are shown
visually at iteration 1, 20, 50, 100, 150, 200,
300 and 500 for
a single run.}
\label{fig:graph_results}
\end{figure}
\subsection{A Graphical Example}
Following a previously
established experiment~\cite{griffiths_steyvers04},
we show the utility of Source-LDA by
visualizing topics created with words that correspond to the pixel
locations in a $5 \times 5$ picture; but we add a key difference.
The original topics are augmented, used to generate
a corpus, and then hidden. Only the non augmented topics
are given as input with the goal of discovering the augmented
topics using the corpus and their original topics.
\subsubsection{Experimental Setup}
We start by creating ten topics with
the vocabulary being the set of pixel locations in
a $5 \times 5$ picture. The vocabulary ($V$) and bag of words
representation of a topic ($T_i$) are defined as:
\vspace{1mm}
\[
  V = \{ xy \mid 0 \leq x < 5 \land 0 \leq y < 5 \}
\]
\[
  T_{i} = 
  \begin{cases}
    xy \mid y = i \land 0 \leq x < 5,& \text{if } 0 \leq i < 5\\
    yx \mid y = i \land 0 \leq x < 5,              & \text{otherwise}
\end{cases}
\]
The topics are shown by Figure 5(a) with the intensity ($I$) of a pixel corresponding to word $w$
in topic $t$ equal to: 
\[
  I(w, t) = Max[5 \times P(w|t), 1]
\]
\indent The representation of topics in this manner
leads to a total of $10$ topics. These original topics are then
augmented by pairing
each topic with a random different topic and
swapping a random word (pixel) that is assigned to each
topic given that the swapped words do not belong
to their original assignments.
Figure \ref{fig:graph_topics}(b) shows the
augmented topics which represent a
$20\%$ augmentation rate between the
original topics. From the set of augmented
topics we generate a 2,000 document
corpus using the generative model of LDA.
Each document consists of $25$ words with
topic assignments drawn from a distribution
sampled from the Dirichlet distribution 
parameterized by $\alpha = 1$.
With the knowledge source consisting solely of
the original non augmented topics we run Source-LDA
on the corpus hoping to discover and properly
label the augmented topics. For comparative
analysis we also run EDA and CTM against the
same data set.\par
\subsubsection{Experimental Results} As shown in Figure \ref{fig:graph_results},
Source-LDA discovers the augmented topics given
the set of original topics.
Not only is Source-LDA able to find the topics
correctly to the augmented distributions used in the
generation of the corpus, but it is also able to
match them to their respective non augmented source distributions.
This simple experiment
highlights a big advantage of Source-LDA; which
is the ability to discover topics that differ
from their respective supervised input set. Other
models such as EDA and CTM are unable to
label the augmented topics correctly due to
the topics containing a word (pixel) not in the original
distribution. The comparative average JS divergence was
$0.012$, $0.138$, and $0.43$ for Source-LDA, EDA, and CTM respectively.\par
\setlength{\textfloatsep}{13pt plus 1.0pt minus 2.0pt}
\begin{figure} [!t]
\centering
\includegraphics[trim={0.0cm 0.63cm 0.0cm 1.55cm},clip,width=1\linewidth]{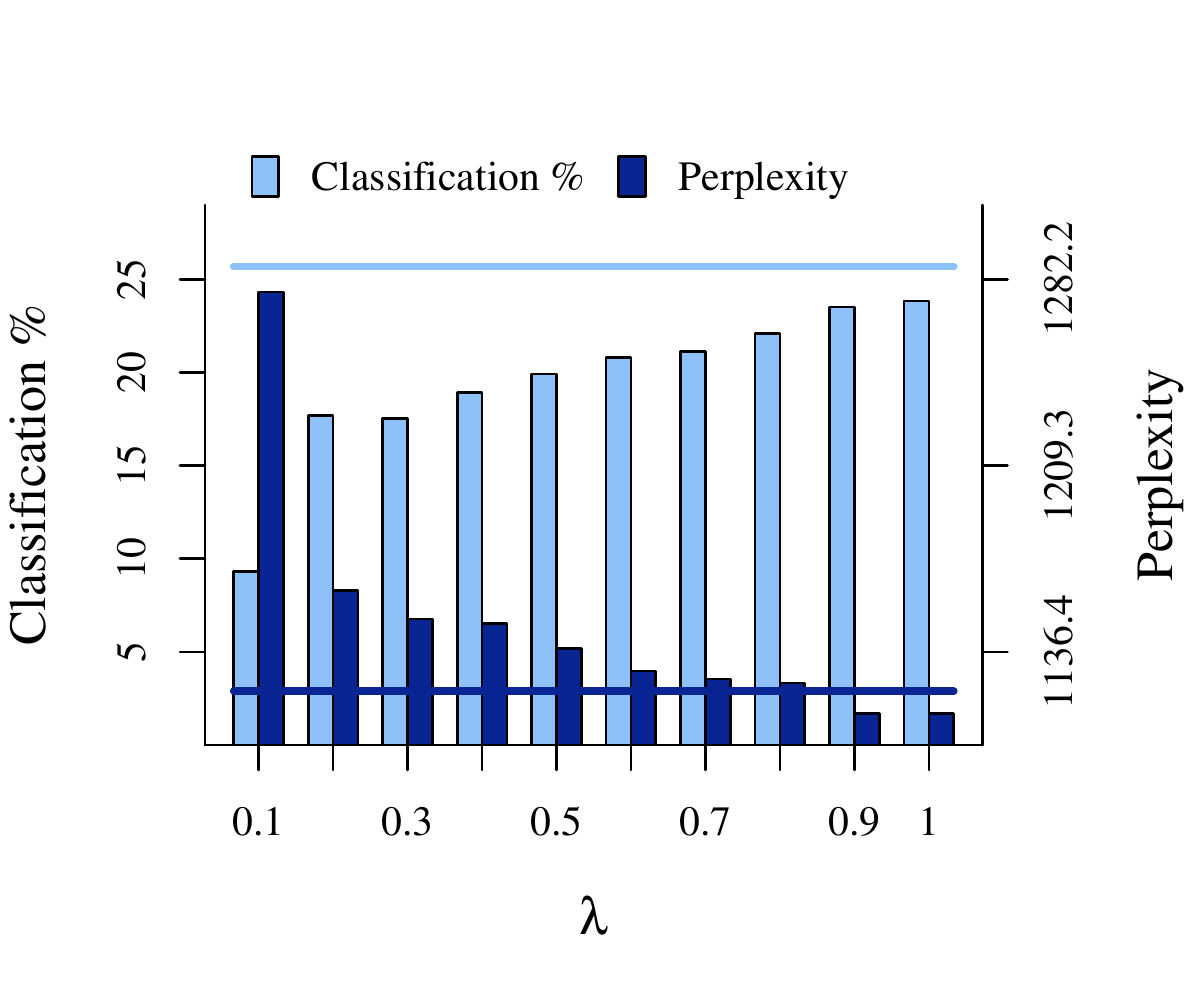}
\caption{Classification accuracy and perplexity
values for fixed values of $\lambda$ compared against the baseline
values generated from a dynamic $\lambda$ with a normal prior. The baseline
values shown as lines represent the classification percentage of
$25.7$ and perplexity value of $1119.9$}
\label{fig:need_results}
\end{figure}
\subsection{Integrating $\lambda$}
A reasonable assumption to a corpus in which
some topics are generated from a knowledge source
is that the topics used in the corpus
are going to deviate (more or less similar) from
their respect source distributions and that each individual
topic is going to deviate at a different rate than other topics.
The introduction
of $\lambda$ to Source-LDA as a parameter to be learned by the
data allows the flexibility of different topics
to be influenced differently by $\lambda$, but
comes at an increase in computation cost.
To show that in certain cases this flexibility
is needed to obtain more accurate results
we derive an experiment consisting of topics
with different deviations from their respective
source distributions. \par
\subsubsection{Experimental Setup}
A synthetic $500$ document corpus is
generated from a 
knowledge source of $100$ randomly selected
Wikipedia topics. The corpus is generated
using the bijective model of Source-LDA as 
outlined in Section 3(A), consisting of
$100$ topics, an average word count per document
of $100$ words, $\mu = 0.5$, $\sigma = 1.0$ and
$\alpha = 0.5$. Furthermore even though for each
topic $\lambda$ was drawn
from $\mathcal{N}(\mu, \sigma^2)$ we bound the
value drawn to the interval $[0,1]$ for comparative
analysis. We then run Source-LDA under the 
bijective model for a baseline of $\mu = 0.5$, $\sigma = 1.0$
against $10$ runs of
Source-LDA with $\lambda$ fixed. After each
run we compare the classification accuracy and
perplexity values.
\subsubsection{Experimental Results}
For all fixed $\lambda$ runs the baseline
approach of varying $\lambda$ in accordance with
the normal distribution results in a higher
classification accuracy. By allowing $\lambda$
to deviate, the model can make up for incorrect
parameter assignments due to a misleading
perplexity value. As shown in Figure \ref{fig:need_results},
classification accuracy is not perfectly correlated
with perplexity. This is shown by the baseline method reporting a higher
perplexity value than the fixed $\lambda = 1$ value while
maintaining a higher classification accuracy. Even though
we still recommend perplexity or
other log-likelihood
maximization approaches to set the parameters
in any unknown data set, maximizing log-likelihood has been
shown to be a less than perfect metric for
evaluating topic models~\cite{tea, corey_ll}. In this experiment
and the remaining experiments we take classification accuracy
to be a more appropriate measurement for evaluating
topic models.
\begin{table*} [htbp]
\centering 
\begin{tabular}{ |l|l|l|l|l|l|l|l|l|}
  \hline
  \multicolumn{3}{|c|}{Inventories} & \multicolumn{3}{c|}{Natural Gas} & \multicolumn{3}{c|}{Balance of Payments} \\
  \hline 
  \multicolumn{1}{|c|}{SRC-LDA}  & \multicolumn{1}{c|}{IR-LDA}  & \multicolumn{1}{c|}{CTM}  & \multicolumn{1}{c|}{SRC-LDA}  & \multicolumn{1}{c|}{IR-LDA}  & \multicolumn{1}{c|}{CTM} & \multicolumn{1}{c|}{SRC-LDA}  & \multicolumn{1}{c|}{IR-LDA}  & \multicolumn{1}{c|}{CTM} \\
  \hline 
  inventory   & systems     & sales     & gas       & corp        & gas           & account     & said    & said             \\
  cost        & products    & year      & natural   & contract    & said          & surplus     & public  & june       \\
  stock       & said        & sold      & used      & company     & total         & deficit     & state   & april                 \\
  accounting  & information & retail    & water     & services    & value         & current     & private & beginning             \\
  goods       & technology  & given     & oil       & unit        & near          & balance     & planned & great            \\
  management  & company     & place     & carbon    & subsidiary  & natural       & currency    & reduce  & later           \\
  time        & data        & marketing & cubic     & completed   & properties    & trade       & local   & remain              \\
  costs       & network     & improved  & energy    & work        & california    & exchange    & added   & reserve              \\
  financial   & kodak       & passed    & fuel      & dlr         & wells         & capital     & make    & equivalent            \\
  process     & available   & addition  & million   & received    & future        & foreign     & did     & imported                  \\
  
  \hline
\end{tabular}
\caption{Topics and their most probable word lists for Source-LDA, IR-LDA, and CTM.}
\label{table:results}
\end{table*}

\subsection{Reuters Newswire Analysis}
To show the type of topics discovered from
Source-LDA we run the model on an existing
dataset. This collection
contains documents from the Reuters newswire from 1987.
The dataset contains 21,578 articles, among a large set of 
categories. One important feature of the dataset are a set
of given categories that we can use for our topic labeling.
These include broad categories such as 
shipping, interest rates, and trade, as well as more refined
categories such as rubber, zinc, and coffee. 
Our choice to apply our topic labeling method to this dataset
is due to the fact that the Reuters dataset is widely used for
information retrieval and text categorization applications.
Due to its widespread use, it can considerably aid us in comparing
our results to other studies. Additionally, because it contains
distinct categories that we can use as our known set of topics,
we can easily demonstrate the viability of our model.
\subsubsection{Experimental Setup}
Source-LDA, LDA, and CTM were run against the 
Reuters-21578 newswire collection. 
Since EDA does not discover new topics,
nor does it update the word distributions of
the input topics, we do not include EDA in
this experiment.
From the original 21,578 document
corpus we select a subset of 2,000 documents.
The Source-LDA and CTM supplementary distributions
were generated by first obtaining a list of topics from the Reuters-21578 dataset. 
Next, for each topic, the corresponding Wikipedia
article was crawled and the words in the topic were counted, forming their
respective distributions. 
Querying Wikipedia resulted in $80$ distinct topics as our superset for the
knowledge source.
Out of the $80$ crawled available topics,
only $49$ topics appear in the 2,000 document
corpus.
This represents the ideal conditions in which Source-LDA is to be applied; that of
a corpus which a significant portion of tokens are generated from a subset of
a larger and relatively easy to obtain topic
set. For all models, a symmetric Dirichlet parameter of $50/T$ (where $T$
is the number of topics) and $200/V$ (where $V$ is the size of the
vocabulary) was used for $\alpha$ and $\beta$ respectively.
For Source-LDA, $\mu$ and $\sigma$ were determined by experimentally
finding a local minimum value of perplexity which
resulted from the parameter values of $0.7$ for
$\mu$ and $0.3$ for $\sigma$.
The bag of words used in the CTM were taken
from the top 10,000 words by frequency for each topic. The models showed good convergence
after 1,000 iterations. After sampling was complete for LDA, the
resulting topic-to-word distribution was mapped using an information
retrieval (IR) approach. The IR approach was to use cosine
similarity of documents mapped to term frequency-inverse document
frequency (TF-IDF) vectors with TF-IDF weighted query vectors formed
from the top 10 words per topic.
\phantom{space space space space space space}
\phantom{space space space space space space}
\\
\indent \textit{2) Experimental Results: }
After the LDA model converged, we label the topics using the IR approach described above (we referred to this topic labeling method as IR-LDA). Given similar labels
from the models it is an intuitive approach to compare the word assignments to each topic model.
Example comparisons are shown in Table \ref{table:results}. The label assignments 
generated from Source-LDA show
a more accurate assignment of labels to
topics than both IR-LDA and CTM. IR-LDA appears to suffer
from mixing of different concepts into a single topic, for
example with the topic ``Inventories,"
the topic assignments could possibly be the combination
of ``Inventories" and ``Information Technology". 
The CTM seems to assign more weight to less important words.
One approach to rectify this problem for CTM is to use a smaller number of
words for the bag of words, but this leads to significant dropout
and no labeled topics are passed through. Out of the total $100$
returned topics, CTM only discovered $6$ labeled topics, with
Source-LDA discovering $15$. Since
the IR approach forces all topics to
a label regardless of the quality of
the label, LDA required all topics
to be matched to a label. Out of the $6$ labeled
CTM topics only $3$ were overlapping with
Source-LDA and IR-LDA and are shown in Table \ref{table:results}. The
remaining $3$ CTM topics were bad matches for the label with
an average of $86\%$ of words not appropriate for the label as
determined by human judgment (we acknowledge the potential
for bias). Meanwhile Source-LDA mismatched
at a rate of $36\%$, with IR-LDA at a rate of $77\%$.  
Source-LDA is more consistent with the meaning of the topic
as opposed to what words you may find when talking about this topic,
which can be generally applied to many concepts.
%
%
%
\subsection{Wikipedia Corpus}
A comparison of Source-LDA against EDA, and CTM
is made using a corpus generated using a known
knowledge source corresponding to medical topics
extracted from MedlinePlus (a consumer-friendly medical dictionary)~\cite{MedlinePlus}.
We evaluate the strength of Source-LDA under
different models proposed in Section 3 using
the metrics of classification
accuracy, JS divergence and Pointwise mutual information (PMI). \par
\indent
 PMI is an established evaluation of
learned topics which takes as input a subset of the most popular tokens comprising a topic
and determines the frequency of all pairs in the subset occurring at a given input distance
from each other in the corpus.
The more that these pairs occur close to each other then the better the learned topics.
PMI differs from  the JS divergence evaluation for this experiment in that
PMI will tell us how good our topics are where as the JS divergence will tell us how good
our distribution over topics for each document is. \par
\subsubsection{Experimental Setup}
A corpus of Wikipedia vocabulary articles was generated by following
the steps of the generative model for Source-LDA, where the chosen $K$ topics
are a subset of a larger collection of Wikipedia topics. 
The topics consisted of $578$ Wikipedia
articles representing the collection of
topic labels from MedlinePlus.
The number
of topics ($K$) was given as $100$, chosen from an entire collection of $578$ topics ($B$),
the number of documents ($D$) was given as $2000$ and the average document word count ($D_{\text{avg}}$) as
$500$, $\mu$ and $\sigma$ were set to $5.0$ and $2.0$ for the bijective evaluation
 $0.7$ and $0.3$ for the Source-LDA model respectively.
 After these $2000$ documents were generated the topic assignments
were recorded and used as the ground truth measurement. The
word assignments were used as the corpus and the different
topic models were applied to these documents. 
The first round of topic models consisted of comparing Source-LDA, EDA, and CTM.
For Source-LDA $\mu$ and $\sigma$ were
set to match that of the generative model.
For all models, a symmetric Dirichlet parameter of $50/T$
and $200/V$
was used for
$\alpha$ and $\beta$ respectively. After
convergence of the models they were
evaluated against the ground truth measurement. In the second round of experiments each topic model was
run under the bijective model, that is they only considered
topics which were used in the ground truth assignments.\par
\indent 
To compare Source-LDA against LDA using PMI, 5 corpora
were generated under the bijective model with
the number of topics $K$ ranging from $100$ to
$200$. $B$, $D$, $D_{\text{avg}}$,
$\mu$, and $\sigma$ were set to $100$, $578$, $200$,
$300$, $1.0$ and $0.0$ respectively. The parameters
for Source-LDA followed the generative model and
all other parameters are the same as the previous experiments. After
$1000$ iterations the top $10$ words given for each topic were used
in the PMI assessment. 
\vspace{.1cm}
\begin {figure*}[!t]
\centering
\subfigure[]  
{
  \includegraphics[trim={0.7cm 1.0cm 0.5cm 0.5cm},clip,width=55mm]{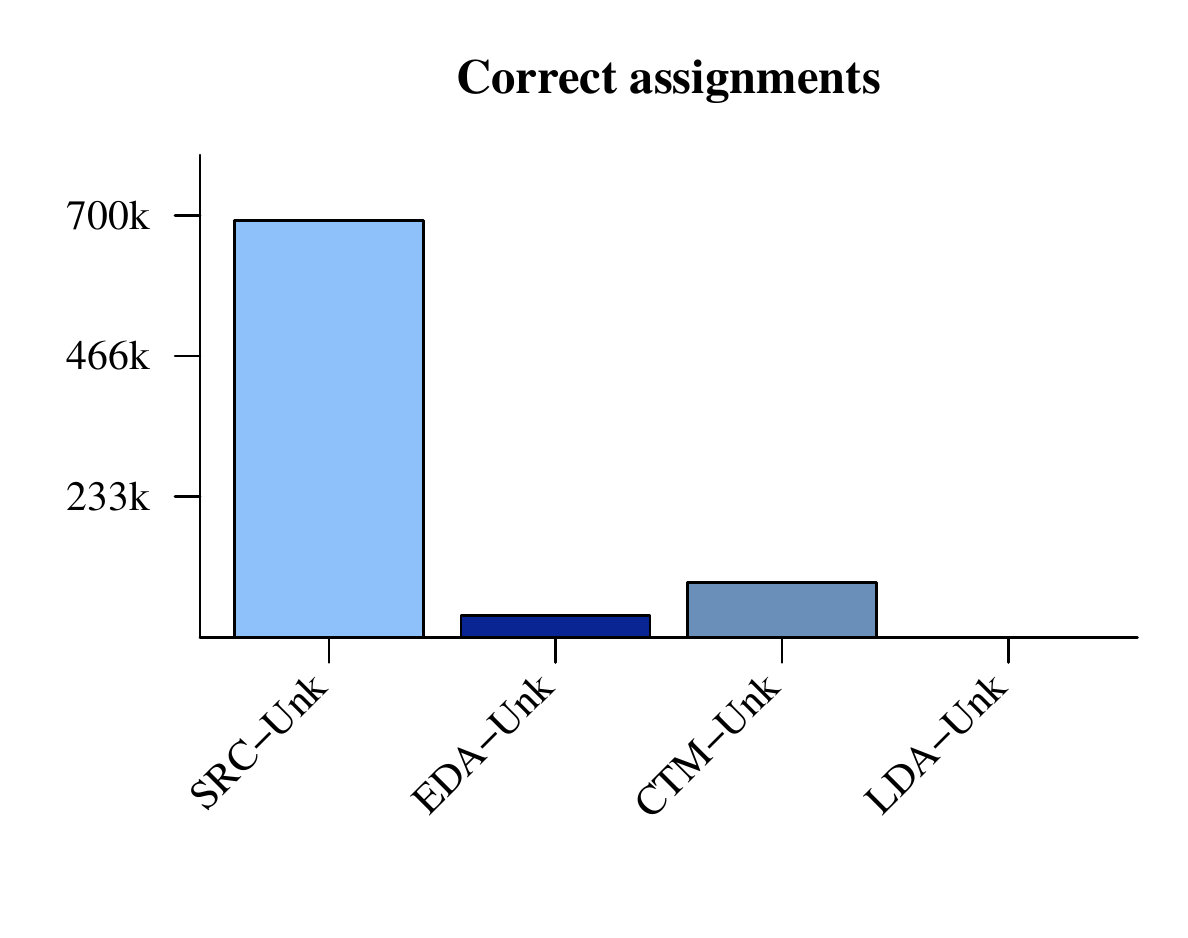}
}
\hfill
\subfigure[]  
{
  \includegraphics[trim={0.7cm 1.0cm 0.5cm 0.5cm},clip,width=55mm]{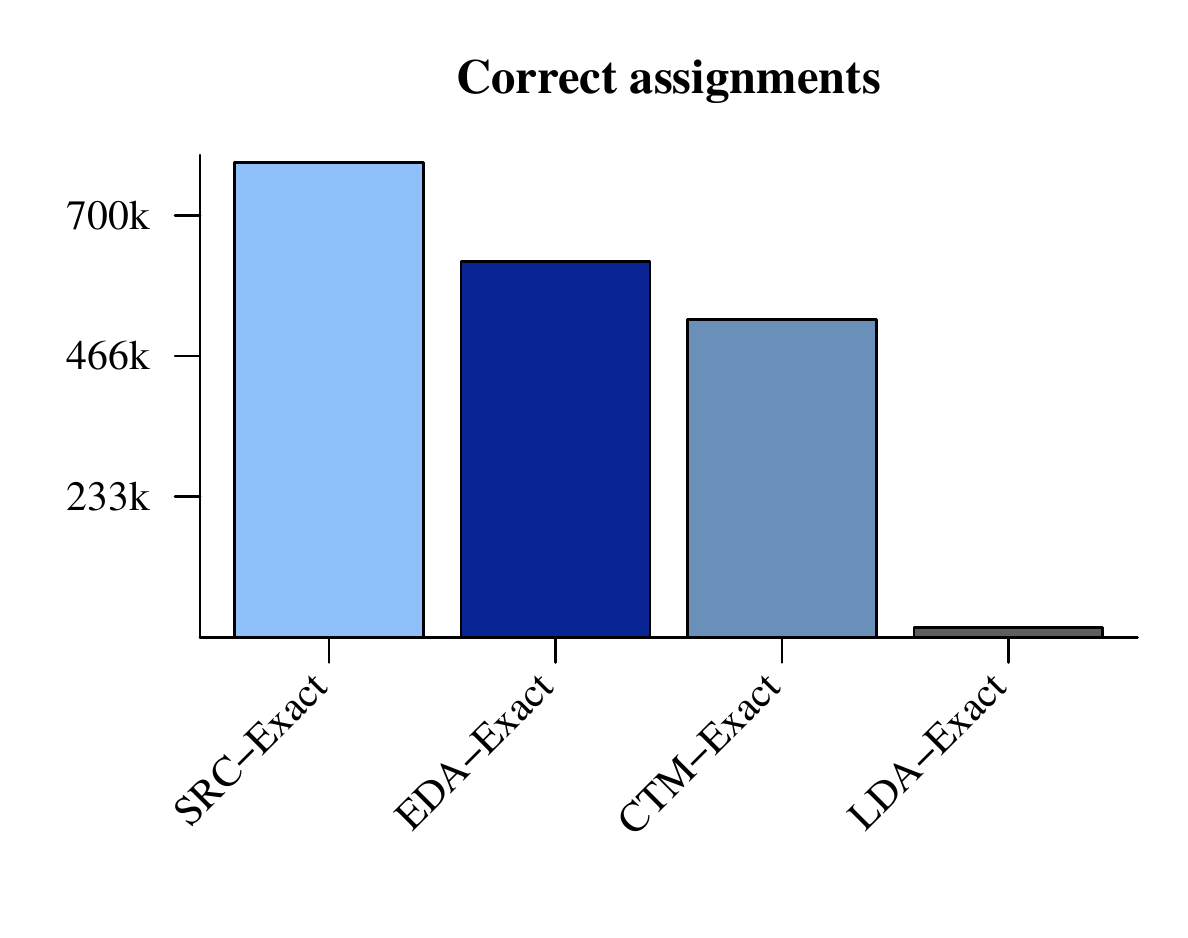}
}
\hfill
\subfigure[]
{
  \includegraphics[trim={0.20cm 0.03cm 0.21cm 0.19cm},clip,width=55mm]{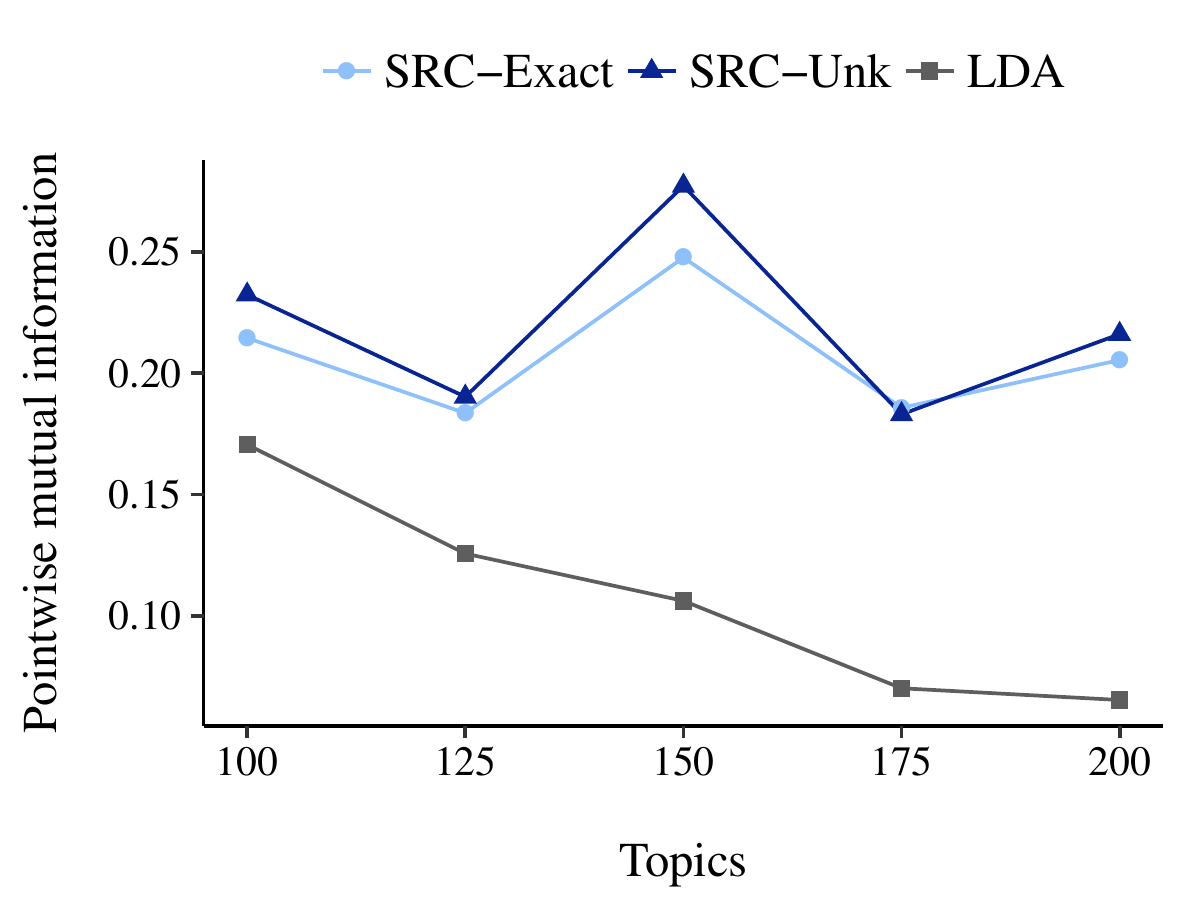}
}
\hfill
\subfigure[]  
{
  \includegraphics[trim={0.7cm 1.0cm 0.5cm 0.5cm},clip,width=55mm]{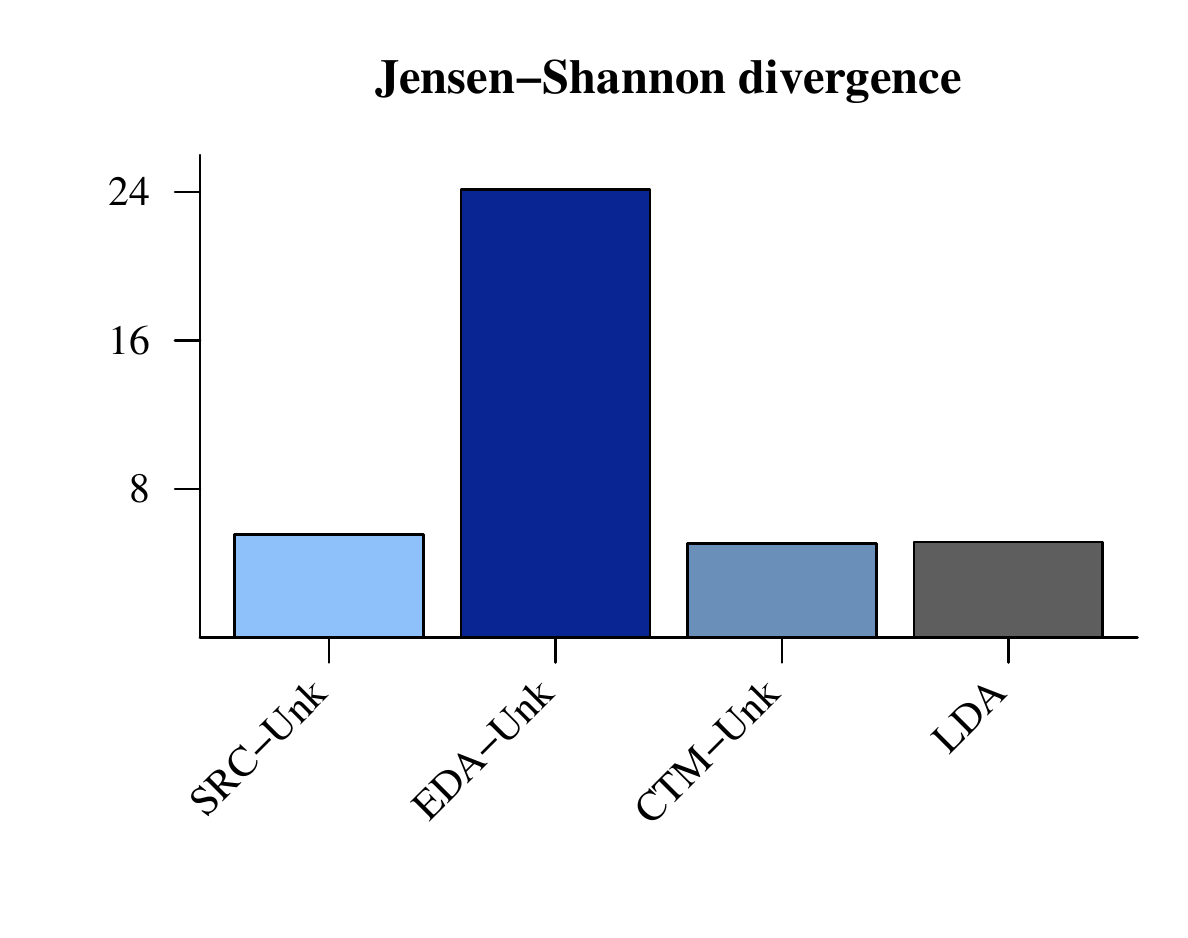}
}
\hfill
\subfigure[]
{
  \includegraphics[trim={0.7cm 1.0cm 0.5cm 0.5cm},clip,width=55mm]{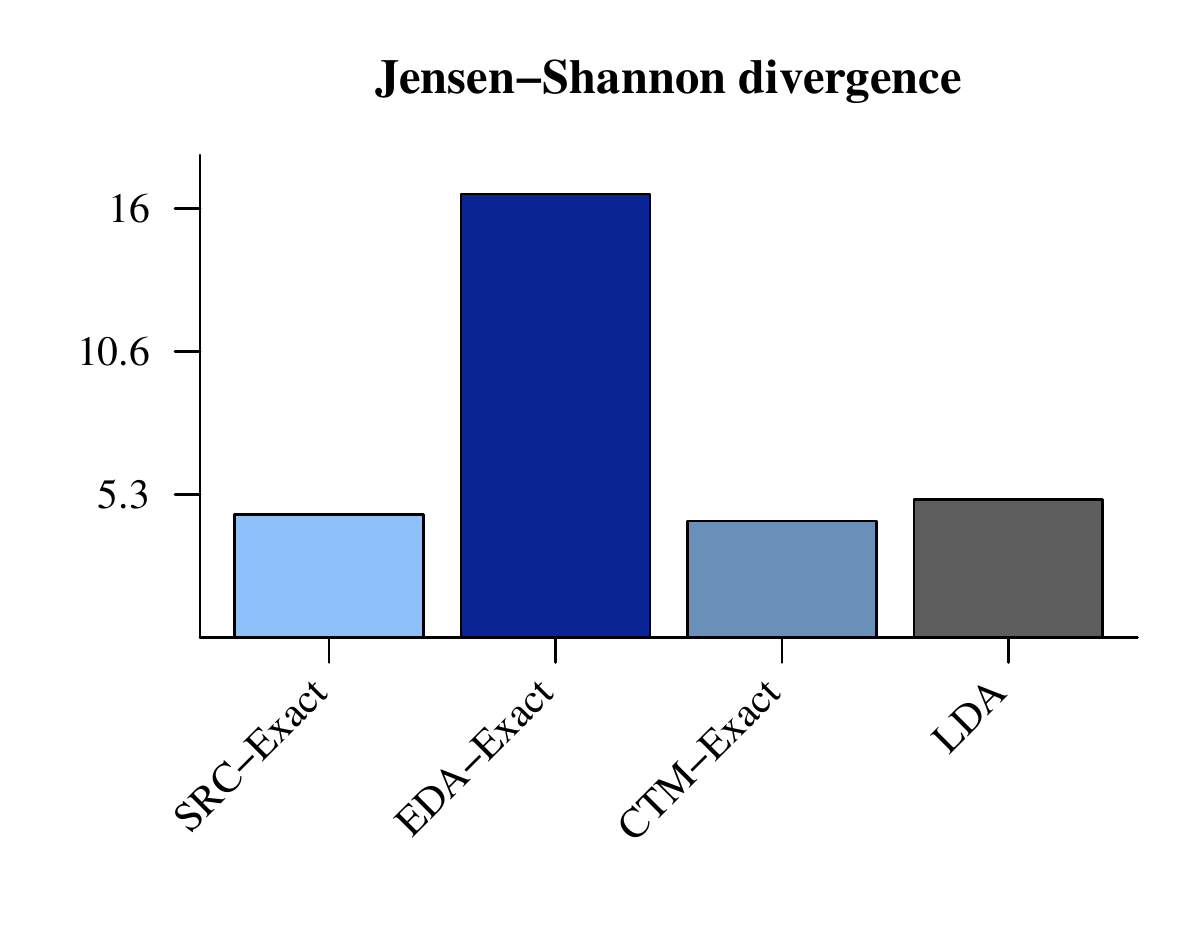}
}
\hfill
\subfigure[]
{
  \includegraphics[trim={0.20cm 0.03cm 0.21cm 0.19cm},clip,width=55mm]{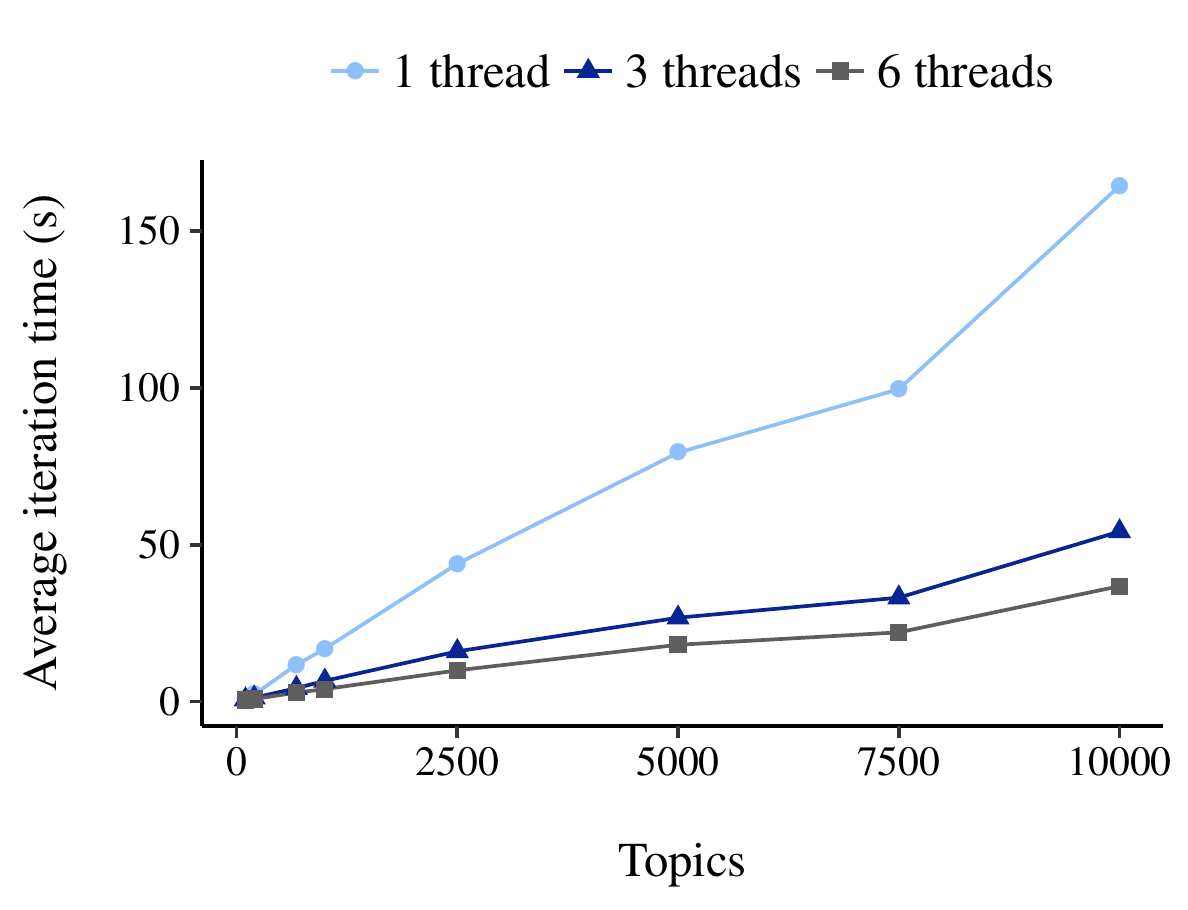}
}
\caption{Results showing the number of
correct topic assignments in the mixed model (a) and bijective model (b) and sum total of the JS divergences of
$\theta$ in the mixed (d) and bijective models (e). Sorted PMI analysis for a Wikipedia
generated corpus inferred by the exact bijective model and mixed model is shown by (c). Performance benchmarking is given in (f). }
\label{results}
\end{figure*}
\subsubsection{Experimental Results}
The topic assignments for each token in the corpus were recorded for all models and the results
compared against each other. Since we know a priori the correct topic assignment for each token we use
the number of correct topic assignments
to be an appropriate measure of
classification accuracy. Note that in
evaluations where the ground truth is
known, classification accuracy is a much
better determination of the goodness of
a model than log likelihood
maximizations such as perplexity and
therefore we do not evaluate the model
using perplexity. In Figure \ref{results}, all topic models run under the full Source-LDA model are tagged with
an ``Unk'' label, and likewise topic models run under the bijective model are tagged with ``Exact''.
The overall number of correct topic
assignments for each model are shown in Figure \ref{results}(a) for
the mixed model and Figure \ref{results}(b) for the bijective model.
Since the LDA model has unknown topics,
JS divergence was used to map each LDA topic to its best
matching Wikipedia topic.
As expected the Source-LDA model (SRC-Unk and SRC-Exact) had the best results amongst all
other topic models for classification accuracy.\par
\indent In the second
analysis  the topic to document
distributions
were analyzed using sorted JS Divergence, and
is irrespective to any unknown mapping.
The results again show
the Source-LDA model to be effective
in accurately mapping topics to
documents whether or not the topics
used in the generative model
are unknown (Figure \ref{results}(d)) or
a known set of topics as shown in Figure \ref{results}(e). 
Even though an accurate alignment of $\theta$ by itself
does not lend much weight to any one model being
superior, we do find it important to demonstrate how $\theta$ is
being affected by the different algorithms.\par
\indent The PMI analysis detailed by Figure \ref{results}(c) show that by PMI, Source-LDA provides a 
better mapping of labels to topics
over the input corpora. This is an encouraging result, even though
the differences are not large, since LDA is a function of topic proximity in a document and word frequency in a topic,
whereas Source-LDA is a function of the same plus the likelihood of a word being in an augmented source distribution. 
\subsection{Performance Benchmarking}
To show the performance gains used by the parallel sampling
algorithm and experiment was set up to generate topics randomly
from a given vocabulary. The corpus was generated using the
same parameters as in Section 4(B) but with $B$ ranging from
$100$ to $10000$. The benchmarking is visualized by Figure \ref{results}(d). It clearly demonstrates that Source-LDA is linearly scalable and easily parallelized. 

\section{Related Work}\label{rel}
Much existing literature exists related to the proposed
approach in this paper. These methods are mainly extensions
of LDA, and add to the original model by introducing enhancements
such as topic labeling, integration with contextual
information and hierarchical modeling. \par
\subsection{Topic Labeling}
In the early research stage, labels were often generated by hand\cite{_abstractautomatic,_abstracta,Mei05discoveringevolutionary,Mei_abstracta}.
Though manual labeling may generate more
understandable and accurate semantics of a topic, it costs a lot of
human effort and it is prone to subjectivity~\cite{Wang06topicsover}. For example, in
the most conventional LDA model, topics are interpreted by
selecting the top words in the distribution~\cite{Blei03latentdirichlet,Lau_automaticlabelling,Wang06topicsover,_abstractautomatic}.
The Topics over Time (TOT) model implements continuous time stamps
with each topic~\cite{Wang06topicsover}. The model has been applied in three kinds of
datasets, and results show more accurate topics and better
timestamp predictions. However, the interpretation of topics is
manual and post-hoc labeling can be time-consuming and
subjective. \par
\indent
Mei et al. proposed probabilistic approaches to automatically
interpreting multinomial topic models objectively.
The intuition of this algorithm was to minimize the semantic
distance between the topic model and the label. To this end,
they extracted candidate labels from noun phrases chunked by an
NLP Chunker and most significant 2-grams. Then they ranked labels
to minimize Kullback-Leibler divergence and maximize mutual
information between a topic model and a label. The approach
achieved the automatic interpretation of topics, but available
candidate labels were limited to phrases inside documents. \par
%
%
%
%
%
\indent
Lau et al came up with an automatic topic label generation method
which obtains candidate labels from Wikipedia articles
containing the top-ranking topic terms, top-ranked document
titles, and sub-phrases. To rank those candidates topic labels,
they used different lexical measurements, such as
point-wise mutual information, Student's t-test,
Dice's coefficient and the log likelihood ratio~\cite{phdthesis}.
Supervised methods like support vector regression
were also applied in the ranking process.
Results showed that supervised algorithm outperforms unsupervised
baseline in all four corpora.\par
\indent
In previous approaches, topics were treated individually and
relation among topics was not considered. Mao et al created
hierarchical descriptor for topics, and results proved that
inner-topic relation could increase the accuracy of topic
labels~\cite{maoautohi}. Hulpus et al proposed a graph-based approach for
topic labeling~\cite{unsuplabel}. In Yashar Mehdad's work, they built an
entailment graph over phrases. Based on that, they then aggregated
relevant phrases by generalization and merging~\cite{MehdadCNJ13}.\par
\indent
Conceptual labeling is an approach to generate a minimum sized set of labels
that best describe a bag of words which includes topics generated from topic
modeling~\cite{concept_labeling}. Concepts used in the topic labeling are taken
from a semantic network and deemed appropriate using the metric Minimum Description Length.
This approach is applied after topic modeling and represents an effective way of labeling topics
over existing approaches.\par
\subsection{Supervised Labeling}
Supervised Latent Dirichlet Allocation (sLDA) is a supervised
approach to labeling topics~\cite{slda}. The approach includes a 
response variable into the LDA model to obtain latent
topics that potentially provide an optimal prediction
for the response variable of a new unlabeled document.
This approach requires, during training, the manual input
of individual topic labels and is constrained to permitting
one label per topic. \par
\indent Similar to sLDA is Discriminative LDA (DiscLDA) which
attempts to solve the same problem as sLDA, but differs in
the approach~\cite{disclda}. The differing approach was centered around
introducing a class-dependent linear transformation on the
topic mixture proportions. This transformation matrix
was learned through a conditional likelihood criterion.
This method has the benefit of both reducing the dimension
of documents in the corpus and labeling the lower dimension
documents. \par
\indent Both sLDA and DiscLDA only allow for a supervised
input set that label a single topic. An approach that
allows for multiple labels in a topic is given by
Labeled LDA (L-LDA)~\cite{labeledlda}. This model differs in the
generation of multinomial distribution theta over the
topics in the model. The scaling parameter is then
modified by a label projection matrix to restrict the
distribution to those topics considered most relevant
to the document. \par
\subsection{Contextual Integration}
An existing approach that takes into account concepts supplied by prior sources
requires a manual input set of relevant terms~\cite{combine}. In the topic model then
these concepts are applied to the assignment of topics to a token in a document. 
Alongside this concept topic modeling a hierarchical
method can also be used to incorporate concepts into
a hierarchical structure. This work shows the utility of bringing in
prior knowledge into topic modeling. \par
\indent An approach that integrates Wikipedia information
into the topic modeling differs than the supervised approach by only requiring
an existing Wikipedia article~\cite{explicit}. The assumption in this work
is that in the generative process
the topics are selected from the Wikipedia word distributions.
The results show that Wikipedia articles can be used
as effective topics in topic modeling. \par
\indent Wikipedia again was shown as a basis for topic
modeling, albeit for a tangential approach, entity 
disambiguation~\cite{explicit}. The approach involved topic modeling as a way of
annotating entities in text. This involved the use of a large dataset
of topics so efficient methods were introduced. Experiments
against a public dataset resulted in a state of the art
performance.
\section{Conclusion}\label{dis}
We have described in this paper a novel methodology for semi-supervised topic modeling with meaningful labels, as
well as provided parallel algorithms to speed up the inference process.
This methodology uses prior knowledge sources to influence a topic model in order to allow
the labels from these external sources to be used for topics generated over a corpus of interest.
In addition, this approach results in more meaningful topics generated based on the quality of the
external knowledge source. We have tested our methodology against the
Reuters-21578 newswire collection corpus for labeling and
Wikipedia as external knowledge sources. The analysis of the quality of topic models
using PMI show the ability of Source-LDA to enhance existing topic models.
\section*{Acknowledgments}\label{ack}
This work was supported by
NIH-NCI National Cancer Institute
T32CA201160 to JW,
the NIH-National Library of Medicine
R21LM011937 to CA, and NIH U01HG008488,
NIH R01GM115833, NIH U54GM114833,
and NSF IIS-1313606 to WW.
The content is solely the
responsibility of the authors and does
not necessarily represent the official
views of the National Institutes of
Health. We would also like to thank
Tianran Zhang, Jiayun Li,
Karthik Sarma, Mahati Kumar, Sara Melvin,
Jie Yu, Nicholas Matiasz, Ariyam Das and all the reviewers for their
thoughtful input into different
aspects of this paper.

\bibliographystyle{IEEEtran}
\bibliography{srcLDA}

\begin{thebibliography}{10}
\providecommand{\url}[1]{#1}
\csname url@samestyle\endcsname
\providecommand{\newblock}{\relax}
\providecommand{\bibinfo}[2]{#2}
\providecommand{\BIBentrySTDinterwordspacing}{\spaceskip=0pt\relax}
\providecommand{\BIBentryALTinterwordstretchfactor}{4}
\providecommand{\BIBentryALTinterwordspacing}{\spaceskip=\fontdimen2\font plus
\BIBentryALTinterwordstretchfactor\fontdimen3\font minus
  \fontdimen4\font\relax}
\providecommand{\BIBforeignlanguage}[2]{{%
\expandafter\ifx\csname l@#1\endcsname\relax
\typeout{** WARNING: IEEEtran.bst: No hyphenation pattern has been}%
\typeout{** loaded for the language `#1'. Using the pattern for}%
\typeout{** the default language instead.}%
\else
\language=\csname l@#1\endcsname
\fi
#2}}
\providecommand{\BIBdecl}{\relax}
\BIBdecl

\bibitem{Blei03latentdirichlet}
D.~M. Blei \emph{et~al.}, ``Latent dirichlet allocation,'' \emph{Journal of
  Machine Learning Research}, vol.~3, pp. 993--1022, 2003.

\bibitem{emr_time}
R.~S. Margalit \emph{et~al.}, ``Electronic medical record use and
  physician-patient communication: an observational study of {Israel}i primary
  care encounters,'' \emph{Patient Education and Counseling}, vol.~1, pp.
  131--141, 2006.

\bibitem{arnold2010clinical}
C.~W. Arnold \emph{et~al.}, ``Clinical case-based retrieval using latent topic
  analysis,'' \emph{AMIA Annual Symposium Proceedings}, vol. 2010, p.~26, 2010.

\bibitem{BisginLFXT11}
H.~Bisgin \emph{et~al.}, ``Mining {FDA} drug labels using an unsupervised
  learning technique - topic modeling,'' \emph{{BMC} Bioinformatics}, vol.~12,
  no. {S-10}, p. S11, 2011.

\bibitem{corey_meta}
\BIBentryALTinterwordspacing
W.~Speier, M.~K. Ong, and C.~W. Arnold, ``Using phrases and document metadata
  to improve topic modeling of clinical reports,'' \emph{Journal of Biomedical
  Informatics}, vol.~61, pp. 260--266, 2016. [Online]. Available:
  \url{http://dx.doi.org/10.1016/j.jbi.2016.04.005}
\BIBentrySTDinterwordspacing

\bibitem{concept_topic_modeling}
C.~C. others, ``Text modeling using unsupervised topic models and concept
  hierarchies,'' \emph{CoRR}, vol. abs/0808.0973, 2008.

\bibitem{explicit}
\BIBentryALTinterwordspacing
J.~A. Hansen \emph{et~al.}, ``Probabilistic explicit topic modeling using
  wikipedia,'' in \emph{Language Processing and Knowledge in the Web -
  25\textsuperscript{th} International Conference, {GSCL} 2013, Darmstadt,
  Germany, September 25-27, 2013. Proceedings}, ser. Lecture Notes in Computer
  Science, I.~Gurevych, C.~Biemann, and T.~Zesch, Eds., vol. 8105.\hskip 1em
  plus 0.5em minus 0.4em\relax Springer, 2013, pp. 69--82. [Online]. Available:
  \url{http://dx.doi.org/10.1007/978-3-642-40722-2}
\BIBentrySTDinterwordspacing

\bibitem{guidedlda}
\BIBentryALTinterwordspacing
J.~Jagarlamudi \emph{et~al.}, ``Incorporating lexical priors into topic
  models,'' in \emph{{EACL} 2012, 13\textsuperscript{th} Conference of the
  European Chapter of the Association for Computational Linguistics, Avignon,
  France, April 23-27, 2012}, W.~Daelemans, M.~Lapata, and L.~M{\`{a}}rquez,
  Eds.\hskip 1em plus 0.5em minus 0.4em\relax The Association for Computer
  Linguistics, 2012, pp. 204--213. [Online]. Available:
  \url{http://aclweb.org/anthology-new/E/E12/}
\BIBentrySTDinterwordspacing

\bibitem{Minka00bayesianinference}
T.~P. Minka, ``{Bayes}ian inference, entropy, and the multinomial
  distribution,'' 2000.

\bibitem{griffiths_steyvers04}
T.~L. Griffiths and M.~Steyvers, ``Finding scientific topics,''
  \emph{Proceedings of the National Academy of Sciences}, vol. 101, no. Suppl.
  1, pp. 5228--5235, Apr. 2004.

\bibitem{darling2011theoretical}
W.~M. Darling, ``A theoretical and practical implementation tutorial on topic
  modeling and gibbs sampling,'' in \emph{Proceedings of the
  49\textsuperscript{th} annual meeting of the association for computational
  linguistics: Human language technologies}, 2011, pp. 642--647.

\bibitem{griffiths2002gibbs}
T.~Griffiths, ``Gibbs sampling in the generative model of latent dirichlet
  allocation,'' 2002.

\bibitem{github}
M.~W. Beck, ``Average dissertation and thesis length,''
  \url{https://github.com/fawda123/diss\_proc}, 2014.

\bibitem{slda}
\BIBentryALTinterwordspacing
D.~M. Blei and J.~D. McAuliffe, ``Supervised topic models,'' in \emph{Advances
  in Neural Information Processing Systems 20, Proceedings of the Twenty-First
  Annual Conference on Neural Information Processing Systems, Vancouver,
  British Columbia, Canada, December 3-6, 2007}, J.~C. Platt, D.~Koller,
  Y.~Singer, and S.~T. Roweis, Eds.\hskip 1em plus 0.5em minus 0.4em\relax
  Curran Associates, Inc., 2007, pp. 121--128. [Online]. Available:
  \url{http://papers.nips.cc/book/advances-in-neural-information-processing-systems-20-2007}
\BIBentrySTDinterwordspacing

\bibitem{disclda}
\BIBentryALTinterwordspacing
S.~Lacoste{-}Julien \emph{et~al.}, ``Disclda: Discriminative learning for
  dimensionality reduction and classification,'' in \emph{Advances in Neural
  Information Processing Systems 21, Proceedings of the Twenty-Second Annual
  Conference on Neural Information Processing Systems, Vancouver, British
  Columbia, Canada, December 8-11, 2008}, D.~Koller, D.~Schuurmans, Y.~Bengio,
  and L.~Bottou, Eds.\hskip 1em plus 0.5em minus 0.4em\relax Curran Associates,
  Inc., 2008, pp. 897--904. [Online]. Available:
  \url{http://papers.nips.cc/book/advances-in-neural-information-processing-systems-21-2008}
\BIBentrySTDinterwordspacing

\bibitem{labeledlda}
D.~Ramage \emph{et~al.}, ``Labeled {LDA: A} supervised topic model for credit
  attribution in multi-labeled corpora,'' in \emph{Proceedings of the 2009
  Conference on Empirical Methods in Natural Language Processing, {EMNLP} 2009,
  6-7 August 2009, Singapore, {A} meeting of SIGDAT, a Special Interest Group
  of the {ACL}}.\hskip 1em plus 0.5em minus 0.4em\relax {ACL}, 2009, pp.
  248--256.

\bibitem{plda}
\BIBentryALTinterwordspacing
Y.~Wang \emph{et~al.}, ``{PLDA:} parallel latent dirichlet allocation for
  large-scale applications,'' in \emph{Algorithmic Aspects in Information and
  Management, 5\textsuperscript{th} International Conference, {AAIM} 2009, San
  Francisco, CA, USA, June 15-17, 2009. Proceedings}, ser. Lecture Notes in
  Computer Science, A.~V. Goldberg and Y.~Zhou, Eds., vol. 5564.\hskip 1em plus
  0.5em minus 0.4em\relax Springer, 2009, pp. 301--314. [Online]. Available:
  \url{http://dx.doi.org/10.1007/978-3-642-02158-9}
\BIBentrySTDinterwordspacing

\bibitem{adlda}
\BIBentryALTinterwordspacing
D.~Newman \emph{et~al.}, ``Distributed inference for latent dirichlet
  allocation,'' in \emph{Advances in Neural Information Processing Systems 20,
  Proceedings of the Twenty-First Annual Conference on Neural Information
  Processing Systems, Vancouver, British Columbia, Canada, December 3-6, 2007},
  J.~C. Platt, D.~Koller, Y.~Singer, and S.~T. Roweis, Eds.\hskip 1em plus
  0.5em minus 0.4em\relax Curran Associates, Inc., 2007, pp. 1081--1088.
  [Online]. Available:
  \url{http://papers.nips.cc/book/advances-in-neural-information-processing-systems-20-2007}
\BIBentrySTDinterwordspacing

\bibitem{fastlda}
I.~Porteous \emph{et~al.}, ``Fast collapsed gibbs sampling for latent dirichlet
  allocation,'' in \emph{Proceedings of the 14\textsuperscript{th} {ACM}
  {SIGKDD} International Conference on Knowledge Discovery and Data Mining, Las
  Vegas, Nevada, USA, August 24-27, 2008}, Y.~Li, B.~Liu, and S.~Sarawagi,
  Eds.\hskip 1em plus 0.5em minus 0.4em\relax {ACM}, 2008, pp. 569--577.

\bibitem{Blelloch90prefixsums}
G.~E. Blelloch, ``Prefix sums and their applications,'' Synthesis of Parallel
  Algorithms, Tech. Rep., 1990.

\bibitem{wallach}
\BIBentryALTinterwordspacing
H.~M. Wallach \emph{et~al.}, ``Rethinking {LDA:} why priors matter,'' in
  \emph{Advances in Neural Information Processing Systems 22: 23rd Annual
  Conference on Neural Information Processing Systems 2009. Proceedings of a
  meeting held 7-10 December 2009, Vancouver, British Columbia, Canada.},
  Y.~Bengio, D.~Schuurmans, J.~D. Lafferty, C.~K.~I. Williams, and A.~Culotta,
  Eds.\hskip 1em plus 0.5em minus 0.4em\relax Curran Associates, Inc., 2009,
  pp. 1973--1981. [Online]. Available:
  \url{http://papers.nips.cc/paper/3854-rethinking-lda-why-priors-matter}
\BIBentrySTDinterwordspacing

\bibitem{imp_2}
\BIBentryALTinterwordspacing
H.~M. Wallach \emph{et~al.}, ``Evaluation methods for topic models,'' in
  \emph{Proceedings of the 26th Annual International Conference on Machine
  Learning, {ICML} 2009, Montreal, Quebec, Canada, June 14-18, 2009}, ser.
  {ACM} International Conference Proceeding Series, A.~P. Danyluk, L.~Bottou,
  and M.~L. Littman, Eds., vol. 382.\hskip 1em plus 0.5em minus 0.4em\relax
  {ACM}, 2009, pp. 1105--1112. [Online]. Available:
  \url{http://doi.acm.org/10.1145/1553374.1553515}
\BIBentrySTDinterwordspacing

\bibitem{hein}
G.~Heinrich, ``Parameter estimation for text analysis,'' \emph{University of
  Leipzig, Tech. Rep}, 2008.

\bibitem{not_used}
\BIBentryALTinterwordspacing
J.~Kang \emph{et~al.}, ``Transfer topic modeling with ease and scalability,''
  in \emph{Proceedings of the Twelfth {SIAM} International Conference on Data
  Mining, Anaheim, California, USA, April 26-28, 2012.}\hskip 1em plus 0.5em
  minus 0.4em\relax {SIAM} / Omnipress, 2012, pp. 564--575. [Online].
  Available: \url{http://dx.doi.org/10.1137/1.9781611972825.49}
\BIBentrySTDinterwordspacing

\bibitem{tea}
\BIBentryALTinterwordspacing
J.~Chang \emph{et~al.}, ``Reading tea leaves: How humans interpret topic
  models,'' in \emph{Advances in Neural Information Processing Systems 22: 23rd
  Annual Conference on Neural Information Processing Systems 2009. Proceedings
  of a meeting held 7-10 December 2009, Vancouver, British Columbia, Canada.},
  Y.~Bengio, D.~Schuurmans, J.~D. Lafferty, C.~K.~I. Williams, and A.~Culotta,
  Eds.\hskip 1em plus 0.5em minus 0.4em\relax Curran Associates, Inc., 2009,
  pp. 288--296. [Online]. Available:
  \url{http://papers.nips.cc/paper/3700-reading-tea-leaves-how-humans-interpret-topic-models}
\BIBentrySTDinterwordspacing

\bibitem{corey_ll}
\BIBentryALTinterwordspacing
C.~W. Arnold, A.~Oh, S.~Chen, and W.~Speier, ``Evaluating topic model
  interpretability from a primary care physician perspective,'' \emph{Computer
  Methods and Programs in Biomedicine}, vol. 124, pp. 67--75, 2016. [Online].
  Available: \url{http://dx.doi.org/10.1016/j.cmpb.2015.10.014}
\BIBentrySTDinterwordspacing

\bibitem{MedlinePlus}
``Medlineplus [internet],'' \url{https://www.nlm.nih.gov/medlineplus/}.

\bibitem{_abstractautomatic}
Q.~Mei \emph{et~al.}, ``Automatic labeling of multinomial topic models,'' in
  \emph{Proceedings of the 13\textsuperscript{th} {ACM} {SIGKDD} International
  Conference on Knowledge Discovery and Data Mining, San Jose, California, USA,
  August 12-15, 2007}, P.~Berkhin, R.~Caruana, and X.~Wu, Eds.\hskip 1em plus
  0.5em minus 0.4em\relax {ACM}, 2007, pp. 490--499.

\bibitem{_abstracta}
Q.~Mei \emph{et~al.}, ``A probabilistic approach to spatiotemporal theme
  pattern mining on weblogs,'' in \emph{Proceedings of the
  15\textsuperscript{th} international conference on World Wide Web, {WWW}
  2006, Edinburgh, Scotland, UK, May 23-26, 2006}, L.~Carr, D.~D. Roure,
  A.~Iyengar, C.~A. Goble, and M.~Dahlin, Eds.\hskip 1em plus 0.5em minus
  0.4em\relax {ACM}, 2006, pp. 533--542.

\bibitem{Mei05discoveringevolutionary}
Q.~Mei and C.~Zhai, ``Discovering evolutionary theme patterns from text: an
  exploration of temporal text mining,'' in \emph{Proceedings of the Eleventh
  {ACM} {SIGKDD} International Conference on Knowledge Discovery and Data
  Mining, Chicago, Illinois, USA, August 21-24, 2005}, R.~Grossman, R.~J.
  Bayardo, and K.~P. Bennett, Eds.\hskip 1em plus 0.5em minus 0.4em\relax
  {ACM}, 2005, pp. 198--207.

\bibitem{Mei_abstracta}
Q.~Mei and C.~Zhai, ``A mixture model for contextual text mining,'' in
  \emph{Proceedings of the Twelfth {ACM} {SIGKDD} International Conference on
  Knowledge Discovery and Data Mining, Philadelphia, PA, USA, August 20-23,
  2006}, T.~Eliassi{-}Rad, L.~H. Ungar, M.~Craven, and D.~Gunopulos, Eds.\hskip
  1em plus 0.5em minus 0.4em\relax {ACM}, 2006, pp. 649--655.

\bibitem{Wang06topicsover}
X.~Wang and A.~McCallum, ``Topics over time: a non-{Markov} continuous-time
  model of topical trends,'' in \emph{Proceedings of the Twelfth {ACM} {SIGKDD}
  International Conference on Knowledge Discovery and Data Mining,
  Philadelphia, PA, USA, August 20-23, 2006}, T.~Eliassi{-}Rad, L.~H. Ungar,
  M.~Craven, and D.~Gunopulos, Eds.\hskip 1em plus 0.5em minus 0.4em\relax
  {ACM}, 2006, pp. 424--433.

\bibitem{Lau_automaticlabelling}
J.~H. Lau \emph{et~al.}, ``Automatic labelling of topic models,'' in \emph{The
  49\textsuperscript{th} Annual Meeting of the Association for Computational
  Linguistics: Human Language Technologies, Proceedings of the Conference,
  19-24 June, 2011, Portland, Oregon, {USA}}, D.~Lin, Y.~Matsumoto, and
  R.~Mihalcea, Eds.\hskip 1em plus 0.5em minus 0.4em\relax The Association for
  Computer Linguistics, 2011, pp. 1536--1545.

\bibitem{phdthesis}
P.~Pecina, ``Lexical association measures and collocation extraction,''
  \emph{Language Resources and Evaluation}, vol.~44, no. 1-2, pp. 137--158,
  2010.

\bibitem{maoautohi}
\BIBentryALTinterwordspacing
X.~Mao \emph{et~al.}, ``Automatic labeling hierarchical topics,'' in
  \emph{21\textsuperscript{st} {ACM} International Conference on Information
  and Knowledge Management, CIKM'12, Maui, HI, USA, October 29 - November 02,
  2012}, X.~Chen, G.~Lebanon, H.~Wang, and M.~J. Zaki, Eds.\hskip 1em plus
  0.5em minus 0.4em\relax {ACM}, 2012, pp. 2383--2386. [Online]. Available:
  \url{http://dl.acm.org/citation.cfm?id=2396761}
\BIBentrySTDinterwordspacing

\bibitem{unsuplabel}
\BIBentryALTinterwordspacing
I.~Hulpus \emph{et~al.}, ``Unsupervised graph-based topic labelling using
  dbpedia,'' in \emph{Sixth {ACM} International Conference on Web Search and
  Data Mining, {WSDM} 2013, Rome, Italy, February 4-8, 2013}, S.~Leonardi,
  A.~Panconesi, P.~Ferragina, and A.~Gionis, Eds.\hskip 1em plus 0.5em minus
  0.4em\relax {ACM}, 2013, pp. 465--474. [Online]. Available:
  \url{http://dl.acm.org/citation.cfm?id=2433396}
\BIBentrySTDinterwordspacing

\bibitem{MehdadCNJ13}
Y.~Mehdad \emph{et~al.}, ``Towards topic labeling with phrase entailment and
  aggregation,'' in \emph{Human Language Technologies: Conference of the North
  American Chapter of the Association of Computational Linguistics,
  Proceedings, June 9-14, 2013, Westin Peachtree Plaza Hotel, Atlanta, Georgia,
  {USA}}, L.~Vanderwende, H.~D. III, and K.~Kirchhoff, Eds.\hskip 1em plus
  0.5em minus 0.4em\relax The Association for Computational Linguistics, 2013,
  pp. 179--189.

\bibitem{concept_labeling}
\BIBentryALTinterwordspacing
X.~Sun \emph{et~al.}, ``On conceptual labeling of a bag of words,'' in
  \emph{Proceedings of the Twenty-Fourth International Joint Conference on
  Artificial Intelligence, {IJCAI} 2015, Buenos Aires, Argentina, July 25-31,
  2015}, Q.~Yang and M.~Wooldridge, Eds.\hskip 1em plus 0.5em minus 0.4em\relax
  {AAAI} Press, 2015, pp. 1326--1332. [Online]. Available:
  \url{http://ijcai.org/Abstract/15/191}
\BIBentrySTDinterwordspacing

\bibitem{combine}
M.~Steyvers \emph{et~al.}, ``Combining background knowledge and learned
  topics,'' \emph{topiCS}, vol.~3, no.~1, pp. 18--47, 2011.

\bibitem{DBLP:conf/nips/2009}
\BIBentryALTinterwordspacing
Y.~Bengio, D.~Schuurmans, J.~D. Lafferty, C.~K.~I. Williams, and A.~Culotta,
  Eds., \emph{Advances in Neural Information Processing Systems 22: 23rd Annual
  Conference on Neural Information Processing Systems 2009. Proceedings of a
  meeting held 7-10 December 2009, Vancouver, British Columbia, Canada}.\hskip
  1em plus 0.5em minus 0.4em\relax Curran Associates, Inc., 2009. [Online].
  Available:
  \url{http://papers.nips.cc/book/advances-in-neural-information-processing-systems-22-2009}
\BIBentrySTDinterwordspacing

\bibitem{DBLP:conf/nips/2007}
\BIBentryALTinterwordspacing
J.~C. Platt, D.~Koller, Y.~Singer, and S.~T. Roweis, Eds., \emph{Advances in
  Neural Information Processing Systems 20, Proceedings of the Twenty-First
  Annual Conference on Neural Information Processing Systems, Vancouver,
  British Columbia, Canada, December 3-6, 2007}.\hskip 1em plus 0.5em minus
  0.4em\relax Curran Associates, Inc., 2008. [Online]. Available:
  \url{http://papers.nips.cc/book/advances-in-neural-information-processing-systems-20-2007}
\BIBentrySTDinterwordspacing

\bibitem{DBLP:conf/kdd/2006}
T.~Eliassi{-}Rad, L.~H. Ungar, M.~Craven, and D.~Gunopulos, Eds.,
  \emph{Proceedings of the Twelfth {ACM} {SIGKDD} International Conference on
  Knowledge Discovery and Data Mining, Philadelphia, PA, USA, August 20-23,
  2006}.\hskip 1em plus 0.5em minus 0.4em\relax {ACM}, 2006.

\end{thebibliography}
\end{document}